\documentclass{article}     % onecolumn (ditto)
\pdfoutput=1
\usepackage[a4paper,width=130mm,top=35mm,bottom=35mm]{geometry}
\usepackage{graphicx}
\usepackage{adjustbox}
\usepackage{textcase,marvosym,cancel,booktabs,colortbl,nicefrac,ifsym}
\usepackage{amsmath,amsfonts, bm, amssymb,tikz,pgfplots,graphicx, colonequals,mathtools, breqn} %graphicx, MnSymbol
\usetikzlibrary{patterns}
\usepackage{booktabs,multirow}%,multicol}
\usepackage{xspace}
\usepackage{natbib}
\usepackage[utf8]{inputenc}
\usepackage{etex}
\usepackage{tabu}
\usepackage{subfigure}
\usetikzlibrary{arrows,shapes,plotmarks,decorations.pathmorphing} % import for ellipse
\usepackage{enumerate}
\usepackage{array}
\usepackage{hyperref}

\graphicspath{{./fig/}}

\newlength{\figwidth}
\newlength{\figheight}
\pgfplotsset{
	kurze Legende/.style={%
		legend image code/.code={
			\draw[##1,line width=1.8pt]
				plot coordinates {
					(0cm,0cm)
					(0.2cm,0cm)
					};%
				}
		}
}

\usepackage{xcolor,pagecolor}

\usepackage[disable]{todonotes}
\usepackage{array}
\newcolumntype{x}[1]{>{\centering\let\newline\\\arraybackslash\hspace{0pt}}p{#1}}

\newcommand{\ValISparse}{V$^{\text{int}}$-S\xspace}
\newcommand{\ValIE}{V$^{\text{int\&ex}}$\xspace}
\newcommand{\ValI}{V$^{\text{int}}$\xspace}

\newsavebox\MBox
\newcommand\Cline[2][red]{{\sbox\MBox{$#2$}%
  \rlap{\usebox\MBox}\color{#1}\rule[-1.2\dp\MBox]{\wd\MBox}{0.2pt}}}
\newcommand{\mmm}[3]{\ensuremath{#2\,\,^{\scriptscriptstyle #3}_{\Cline[gray]{\scriptscriptstyle #1}}}} % min, median, max
\newcommand{\mm}[2]{\ensuremath{\,^{\scriptscriptstyle #2}_{\Cline[gray]{\scriptscriptstyle #1}}}} % min, median, max
\newcommand{\mmmt}[3]{\ensuremath{#2}&\ensuremath{{\ ^{\scriptscriptstyle #3}_{\Cline[gray]{\scriptscriptstyle #1}}}}} % min, median, max
\DeclarePairedDelimiterX{\norm}[1]{\lVert }{\rVert}{#1}
\DeclarePairedDelimiterX{\abs}[1]{\lvert }{\rvert}{#1}
\newcommand{\psiT}{\psi_{\theta} }
\newcommand{\dEQL}{EQL\!$^\div$\xspace} %{\textcolor{green}{EQL}}%{EQL,ABFNet}

\newcommand{\pysr}{$\mathrm{GA}$\xspace}
\newcommand{\pysrA}{$\mathrm{GA1}$\xspace}
\newcommand{\pysrB}{$\mathrm{GA2}$\xspace}
\newcommand{\bEQL}{$\mathrm{iEQL}$\xspace}

\newcommand{\bEQLmotor}{$\mathrm{iEQL_{\text{motor}}}$\xspace}
\newcommand{\eql}{$\mathrm{EQL}$\xspace}

\renewcommand{\cite}[1]{\citet{#1}}

\makeatletter
\DeclareRobustCommand\onedot{\futurelet\@let@token\@onedot}
\def\@onedot{\ifx\@let@token.\else.\null\fi\xspace}
\def\eg{e.g\onedot}
\def\ie{i.e\onedot}

\def\etc{etc\onedot}

\makeatother

\newcommand{\q}{\quad}
\newcommand{\qq}{\qquad}

\newcommand{\Ber}{\mathrm{Ber}}

\newcommand{\g}{\mid}

\newcommand{\e}{\mathrm{e}}

\definecolor{lred}{RGB}{200,0,0}
\definecolor{dred}{RGB}{130,0,0}
\definecolor{dblu}{RGB}{0,0,130}
\definecolor{dgre}{RGB}{0,130,0}
\definecolor{dgra}{RGB}{50,50,50}
\definecolor{mgra}{RGB}{221,222,214}
\definecolor{lgra}{RGB}{238,238,234}

\definecolor{TUred}{RGB}{141,45,57}
\definecolor{TUdark}{RGB}{55,65,74}
\definecolor{TUgold}{RGB}{174,159,109}
\definecolor{TUgray}{RGB}{175,179,183}

\definecolor{bgnote1}{RGB}{254,178,76}
\definecolor{bgnote2}{RGB}{158,202,225}
\definecolor{bgnote3}{RGB}{153,216,201}
\definecolor{notepagebg}{RGB}{230,230,230}

\def\tableref#1{table~\ref{#1}}
\def\Tableref#1{Table~\ref{#1}}
\def\figref#1{figure~\ref{#1}}
\def\Figref#1{Figure~\ref{#1}}

\def\secref#1{section~\ref{#1}}
\def\eqref#1{equation~\ref{#1}}
\def\plaineqref#1{\ref{#1}}
\def\1{\bm{1}}
\newcommand{\train}{\mathcal{D}}

\def\eps{{\epsilon}}

\def\vb{{\bm{b}}}

\def\vg{{\bm{g}}}

\def\vx{{\bm{x}}}
\def\vy{{\bm{y}}}
\def\vz{{\bm{z}}}

\def\mW{{\bm{W}}}
\def\mX{{\bm{X}}}
\def\mY{{\bm{Y}}}

\DeclareMathAlphabet{\mathsfit}{\encodingdefault}{\sfdefault}{m}{sl}
\SetMathAlphabet{\mathsfit}{bold}{\encodingdefault}{\sfdefault}{bx}{n}
\def\gN{{\mathcal{N}}}

\newcommand{\E}{\mathbb{E}}
\newcommand{\Ls}{\mathcal{L}}
\newcommand{\R}{\mathbb{R}}

\newcommand{\reg}{\lambda}

\DeclareMathOperator*{\argmax}{arg\,max}
\DeclareMathOperator*{\argmin}{arg\,min}

\begin{document}

\title{Informed Equation Learning
}

\author{Matthias Werner
        \thanks{corresponding author: matthias.werner@tuebingen.mpg.de}
        \thanks{Machine Learning Group at ETAS GmbH, Bosch Group}
        \thanks{Max Planck Institute for Intelligent Systems, Tübingen, Germany}
        \thanks{University of Tübingen}        \and
        Andrej Junginger\footnotemark[1] \and 
        Philipp Hennig\footnotemark[3]\footnotemark[2] \and 
        Georg Martius\footnotemark[2]}

\maketitle

\begin{abstract}
Distilling data into compact and interpretable analytic equations is
one of the goals of science.
Instead, contemporary supervised machine learning methods mostly produce unstructured and dense maps from input to output.
Particularly in deep learning, this property is owed to the
generic nature of simple standard link functions.
To learn equations rather than maps, standard non-linearities can be replaced with structured building blocks of atomic functions.
However, without strong priors on sparsity and structure, representational complexity and numerical conditioning limit this direct approach.
To scale to realistic settings in science and engineering, we propose an informed equation learning system.
It provides a way to incorporate expert knowledge about what are permitted or prohibited equation components, as well as a domain-dependent structured sparsity prior.
Our system then utilizes a robust method to learn equations with atomic functions exhibiting singularities, as \eg~logarithm and division.
We demonstrate several artificial and real-world experiments from the
engineering domain, in which our system learns interpretable models of high predictive power.
% \keywords{First keyword \and Second keyword \and More}

\end{abstract}

\section{Introduction}\label{sec:introduction}

Mathematical models formed of equations are key elements in natural sciences to
describe phenomena and their underlying principles. %between components.
Physics, in particular, uses equations to build a coherent description of natural laws.
For example, Newton's and Kepler's laws or Schrödinger's equation describe and encode
the phenomena of motion, celestial dynamics~\citep{Feynman1965}, or quantum mechanics~\citep{Ballentine1970}, respectively.
In the engineering domain, mathematical equations are used, e.g.,
in model-predictive control~\citet{GarciaPrettMorari1998:MPC}
or as components describing complex system.
% \philippi{I rephrased this sentence, is this okish?}
The designed equations then correspond to a hypothesis for the inner system's behavior and represent its relations and properties.
Implementing such control strategies into embedded control units for
industrial-scale applications raises various additional constraints like low
computational effort, minimal memory usage, low effort of calibration,
and generalization to similar systems.

Learning mathematical models in an automated fashion is referred to as
\emph{equation learning} or \emph{symbolic regression}.
The desired mathematical expressions are usually compositions of \emph{atomic units} consisting of a set of functions \(\{f_i\}_{i\leq F}\) and a set of operations that connect the single functions.
The former can be, e.g., functions like
$\{\sin, \,\mathrm{sqrt},\, \log,\,\dots\}$
and they can be connected by operations such as
$\{+,-,*,/,\circ,\dots\}$.
Out of these atomic units, equation learning finally tries to infer an analytic expression to describe the relations between the quantities of interest.
In this work, we consider a Gaussian regression setting with a parameterized analytical function $\psiT$, to map a $D$-dimensional input $\vx$ to a $D'$-dimensional output $\vy$
\begin{equation} \label{eq:symbolic_regression}
    \vy_i = \psiT(\vx_i) + \boldsymbol{\eps}\;, \qq  \boldsymbol{\eps} \sim \gN(0,\gamma^2)\;.
\end{equation}
Here, $\theta$ denotes the parameters, the dataset $(\mX, \mY) \in \train$ is assumed to be sampled iid., and it contains $N$ data points each of which
has Gaussian noise $\boldsymbol{\epsilon}$ with variance $\gamma^2$.
The structure of $\psiT$ as well as its parameters shall then be inferred from the data.
% \medskip
Many contemporary supervised machine learning algorithms optimize for prediction performance alone.
Thus, they typically produce unstructured and dense equations, which do not reveal insight into the underlying principles of the data-generating system.
Following the principle of Occam's razor~\citep{MacKay1992},
an equation learning system must trade-off
complexity against expressivity to overcome this issue.
This typically results in a regularization of the objective function.
A generic symbolic regression method using machine learning, named \eql~\citep{Martius2017},
replaces standard link functions in a multi-layer feed-forward neural network
with atomic units ($\sin,\, \cos,\, \mathrm{identity},\,*$)
and a Lasso regularization \citep{Tibshirani1996}.
In principle, this approach can be extended to
all continuously differentiable functions with unbounded domains.
However, for common atomic units, like $\mathrm{division}$
 or $\mathrm{logarithm}$
with a half-bounded domain $\mathbb{D}$
or with a singularity,
continuous optimization of an equation-learning neural network can break down
for two reasons:
First, the cascading transformation of intermediate results in a deep architecture can project values outside of the constrained input domains.
Applying domain-limiting mapping functions, such as $\mathrm{softplus}$,
is not ideal since it compromises the final symbolic equation.
Second, singularities in the atomic functions or their derivative can lead to large gradients and make optimization highly unstable.
A first approach incorporating divisions in the final layer in \eql has been presented in~\citet{Sahoo2018}.
Their algorithm uses a hard-coded curriculum, which depends on the number of epochs itself.
This makes it hard to apply to datasets with different sample sizes.
In contrast, our algorithm overcomes those issues and introduces a learnable parameter instead.
It is a robust and simple algorithm for atomic units with singularities.
We show that this unlocks new classes of functions, as \eg division, logarithm, \etc
in all layers of the \bEQL, which is not covered by previous publications.

In general, a large expressivity of the equation learner is desired.
The \eql architecture is, in principle, well suited for scaling to large problems.
However, the set of possible solutions also grows drastically.
Thus, in general, one cannot expect to obtain a symbolic expression that is easily interpretable or that captures the true correspondences.
Therefore, we propose to incorporate expert knowledge into the system such that suitable solutions are preferred.
Our approach is based on excluding certain combinations of functions or on providing a user-dependent weighting scheme.
It was developed in close cooperation with applied engineers.
Technically, this is achieved by designing a generalized sparsity regularization,
that allows one to choose specific complexities for each atomic unit,
and that can flexibly prohibit certain, user-defined function compositions.

In this paper we advance the state-of-the-art in symbolic regression by
\begin{enumerate}[(a)]
\item
proposing an informed equation learner (\bEQL) allowing to
incorporating expert knowledge,
      \item a robust training method for atomic units with
          singularities (\eg logarithm and division), and
      \item showing that interpretable expressions
        can be discovered from real-world datasets from the engineering domain.
\end{enumerate}
The paper is structured as follows:
First, we discuss related work in section~\ref{sec:related-work}.
In section~\ref{sec:method}, we introduce the architecture of the \bEQL, outline
the instance selection criteria and a domain specific complexity measure
with its sparsity inducing regularization.
Moreover, we present the robust training method for atomic units with
singularities.
In section~\ref{sec:experiments}, we present applications of our method to
artificial datasets as well as two real-world applications in the engineering
domain, and we conclude in section~\ref{sec:conclusion}.
\section{Related Work}
\label{sec:related-work}

Equation learning studies the relation between input and output data.
As a thread within the wider area of explainability, it is of increasing importance to
machine learning, which mostly produces black box models.
A major challenge in equation learning is to efficiently search the space of
possible expressions,
which increases exponentially with the number of atomic units
required to describe the data relation.
Current machine learning methods for equation learning can be separated in bottom up
and top down search strategies.

Bottom up methods improve and learn based on individually sampled mathematical
expressions until the final expression is simple and accurate.
For this purpose, machine learning offers efficient search strategies
~\citep{Zaremba2014,Kusner2017,Li2019,Lample2020}.
A physics-inspired method for equation learning that uses neural networks
to identify symmetries and separability within the data was introduced by \citet{Udrescu2019}.%Udrescu et al.~\cite{Udrescu2019}.
Common methods to overcome the exponential growth of the search space are
genetic programming
and evolutionary
algorithms~%
\citep{Langdon2008,Searson2010,McRee2010}.
Algorithms in this vein have been used to automate the discovery of natural laws~\citep{Dub2011,SchmidtLipson2009}.
However, none of these methods use gradient information about the learned
expressions during training. This was addressed in a reinforcement learning formulation by \citet{Petersen2019}. They transformed
the discrete optimization to a continuous gradient based optimization scheme via
risk-seeking policy gradients, which is a modification of the
\textsc{Reinforce}~\citep{Williams1992} policy gradient.

Top down search strategies are in contrast to these
bottom up search strategies.
They start with a highly general expression and try to omit irrelevant
parts during training until a simple but accurate expression is found.
This idea is also applied by the \eql\citep{Martius2017}, which defines a neural
network architecture as general expression.
It represents all possible symbolic expressions within its architecture.
During training irrelevant parts get pruned until the \eql itself
represents a simple but accurate equation.
A similar architecture has been studied by \citet{Long2019} on % Long et al.~\cite{Long2019} for partial
differential equations, and
the integration of an \eql-like architecture within other deep learning
frameworks has been studied by \citet{Kim2019}. %Kim et al.~\cite{Kim2019}.
Also in physics it has been
used to obtain analytical expressions of classical free energy functionals in~\citet{Oettel2020}.
Generally, these top down methods have the advantage that they permit the
use of efficient continuous gradient based optimization.
In contrast to~\citep{Martius2017,Sahoo2018}, which used $L_1$ regularization,
we use $L_0$ regularization~\citep{Louizos2018}, which does not
shrink the weights during training and allows to naturally align a domain
specific complexity measure with the regularization.
A first approach to use divisions in \eql has been presented by \citet{Sahoo2018}.%Sahoo et al.~\cite{Sahoo2018}.
Their architecture only considers division units in the final layer.
We propose a robust training method for atomic units
with singularities in general that does not require a predefined curriculum
 and allows to using division units in all layers.

\section{Method}
\label{sec:method}

\subsection{Architecture of the \bEQL}
    % \input{sections/eql}

%%%%%%%%%%%%%%%%%%%%%%%%%%%%%%%%%%%%%%%%%%%%%%%%%%%%%%%%%%%%%%%%%%%%%%%%%%%%%%%%
\begin{figure}[tbp]
	\centering
    \input{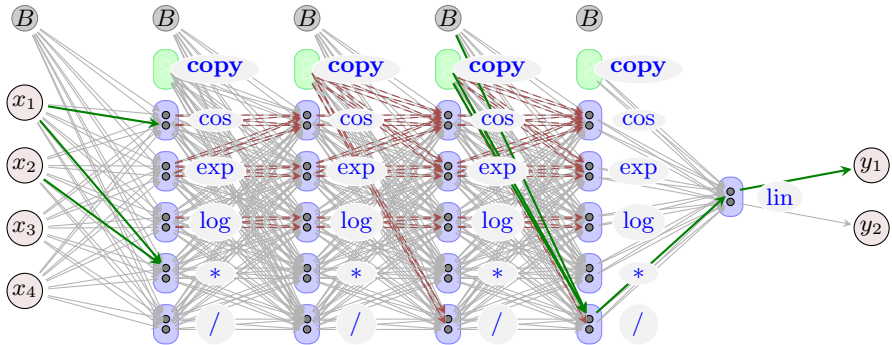}
    \vspace{-.5em}
    \caption{%
        Exemplary setup of the \bEQL with 4 hidden layers, three types of unary atomic units
        \(\{\cos,\,\exp{\;},\,\log \}\) and
        two types of binary atomic units
        \(\{*,\,/\}\)
        per layer.
        The $\mathrm{copy}$ units contains all units from all previous layers.
        The red dashed arrows mark forbidden connections (see \eqref{eq:prohibited_components}) which are removed, also
        from the $\mathrm{copy}$ units.
        After successful training, the \bEQL represents the desired equation.
        The green connections visualize an exemplary model for the equation $y_2 = 0$
        and
        $y_1=(x_2\cdot x_1)/(\cos x_1 +1.5)$.
        }
\label{fig:eql}
\end{figure}
%%%%%%%%%%%%%%%%%%%%%%%%%%%%%%%%%%%%%%%%%%%%%%%%%%%%%%%%%%%%%%%%%%%%%%%%%%%%%%%%
%
Our goal is to identify a compact equation describing data accurately.
A standard neural network as a universal function approximator
is representing a large uninterpretable equation.
Replacing units of a neural network with various building blocks of equations
 and forcing a sparse connection in the network can yield simple and interpretable
 mathematical equations.
The proposed architecture, see \Figref{fig:eql}, of a multi-layered feed-forward equation learning network is similar to
previous work by~\citet{Martius2017,Sahoo2018}.
It consists of $L$ layers, whose $L-1$ hidden layers correspond to the
maximum number of function compositions.
Each hidden layer applies a non-linear transformation, which consists of
$u$ unary atomic units $f_{i\leq u}: \R\to\R$ and $v$ binary atomic units
$g_{j\leq v}: \R\times\R \to \R$.
The latter can be, e.g., division $a/b$ or multiplication $a \cdot b$.
We apply an affine transformation on the output of all preceding layers,
including the input data ($\vy^0 :=\vx$), instead of just the previous one (similarly used in dense-nets \citep{Huang2016})
\begin{align}
    \vz^{l}&=\mW^{l} \tilde\vy^{l-1}+\vb^l\,,\, \text{with: }\q
    \tilde\vy^{l-1}=(\vy^0,\dots,\vy^{l-1})\\
    y^{l} &= \big(f_1(z_1^l),.., f_u(z_u^l),
           g_1(z_{u+1}, z_{u+2}),\dots\big).
\end{align}
Those skip connections allow identified sub-solutions (terms in the equation) to be available with no additional
cost to all subsequent layers.
\Figref{fig:eql} visualizes a possible layout of the architecture, in which
all previous' layers output is marked by the $\mathbf{copy}$ units.
Naturally, the skip connections favor less nested expressions and
allow to ignoring unnecessary layers.
Moreover, the search for appropriate hyperparameters is simplified as
the initial network can be chosen larger and more complex.
Depending on the application, certain combinations of atomic units like
\begin{align}\label{eq:prohibited_components}
    \cos(\cos(\cdot)) &&
    \cos(\exp(\cdot)) &&
    \exp(\exp(\cdot)) &&
    % \nicefrac{a}{\nicefrac{b}{c}} &&
    \log(\log(\cdot)
\end{align}
with respective arguments can be undesirable (\ie~they should only be chosen rarely), or they might not make sense at all.
\emph{Expert} and \emph{domain knowledge} can provide such information on possible equation structures.
Therefore, all forbidden connections are removed from the architecture
as shown in \figref{fig:eql}.
That way, an expert finally decides which combinations to use and which
to exclude.
Such information could not be used in previous work by
\citet{Martius2017,Sahoo2018},
since the respective architecture relied on
$\mathrm{linear}$ units, which accumulate output of all
previous layer in an affine transformation.
That made it impossible to remove certain combinations.
However, with the present framework based on $\mathrm{copy}$ units,
it is easily possible to prohibit certain combinations of atomic units in
the \bEQL.
\subsection{Model Selection Criteria}\label{sec:model_selection}
We apply model selection criteria \ValI, \ValISparse and \ValIE proposed by
\cite{Sahoo2018} to select a good equation amongst all found equations.
They are based on a normalized validation error $\tilde\nu_{\text{in}}$, normalized complexity $\tilde s^2 $
and a normalized extrapolation validation error $\tilde\nu_{\text{ex}}$,
which requires some points (here 40) from the extrapolation domain
\begin{align}\label{eq:selection_criteria}
    \text{\ValI}=\argmin[\tilde\nu_{\text{int}}^2]\\
    \text{\ValISparse}=\argmin[\tilde\nu_{\text{int}}^2+\tilde s^2]\\
    \text{\ValIE}=\argmin[\tilde\nu_{\text{int}}^2+\tilde\nu_{\text{ex}}^2]\,.
\end{align}

\subsection{Complexity Measure}\label{sec:c_measure}

The ``complexity'' of an equation --- not the computational complexity per se, but whether an expert would consider it ``basic'' or ``involved'', depends both on the number of terms and a domain specific complexity cost of each atomic operation.
To capture this aspect, we use a weighted sum of the number of parameters as a measure of complexity.
This is a commonly accepted way to introduce domain knowledge in symbolic regression, \eg in evolutionary search \citep{Dub2011}.
An expert defines domain-specific complexity factors $c_u$ for each atomic unit type $u$. % for the weighting.
The $c_u$ can depend on soft criteria, like domain or personal experience, but also quite concretely on computational cost.
A specific choice of the complexity factors $c_u$ for our experiments is shown in \tableref{ta:complexity_units}.

\begin{table}
    \caption{Domain specific complexity factors $c_u$ for different atomic units.
      No preference is denoted as \textit{plain}.
      For the real world applications (power loss of an electric machine (section~\ref{sec:power_loss})
      and torque model of a combustion engine (section~\ref{sec:comb_engine})).
      We use values suggested by domain experts denoted as \textit{motor}.
    }%\matthias{remove theory of relativity??} }
    \label{ta:complexity_units}
  \centering
   \begin{adjustbox}{max width=\linewidth}
       \begin{tabular}{l c c c c c c c c c}%{l@{\ \ } *{11}{c@{\ \ \ }}}
        \toprule
        & $\nicefrac{+}{-}$ &\  $*$\  &\  $/$\  & $x^2$&$\log$& $\sqrt{\phantom{x}}$&  $\exp$& $\cos$\\% & $\sqrt{1-x^2}$ & $\sinh$   &$\cosh$\\
        \midrule
            plain         & 1     & 1   & 1   &  1   & 1   &   1        &    1    & 1                \\% & --              & --         & -- \\
      %       F         & 1     & 1   & 1   &  1   & 10   &   10        &
      %    10    & 1    & --              & --         & --    \\
             motor         & 1     & 2   & 5   &  2   & 5   &   3        &    5    & 10              \\% & --              & --         & --    \\
             % relativity  & 1     & 2   & 5   &  2   & 10   &   4        &
         % 8     &10     & 1              & 1         & 1 \\
        \bottomrule
    \end{tabular}
    \end{adjustbox}
\end{table}

\subsection{Sparsity Inducing Method}\label{app:l0_method}
Sparsity inducing regularization is necessary to extract interpretable and compact equations from the \bEQL.
We are able to directly optimize for the complexity measure, described in
\secref{sec:c_measure}, as regularization $\Ls_{\text{C}}$
and the mean squared error as data loss $\Ls_{\text{D}}$:
\begin{align}
\Ls_{\text{D}} = \frac{1}{N} \sum\limits_{i=1}^N
\norm{\vy_i-\mathrm{iEQL}(\vx_i, \mW)}_2^2 \,,
\quad
\Ls_{\text{C}} = \sum\limits_{u=1}^{U} c_u \norm{W_u}_0 \,.
\end{align}
$W_u$ represents all weights, which correspond to an atomic unit type.
Opposed to the $L_1$ regularization scheme used by~\citet{Martius2017,Sahoo2018},
we use a differentiable version of $L_0$-regularization~\citep{Louizos2018}.
A final retraining of the pruned network is not necessary anymore, as it does not
induce shrinkage on the actual values of the weights.
Minor modifications are required to apply the original implementation of~\citet{Louizos2018}
to weights instead of nodes.
Each weight is multiplied by a non-negative stochastic Bernoulli distributed gate $g_j\sim\Ber(\pi_j)$.
The dropout rate of each weight is thus given by $1-\pi_j$.
The expectation of a weight being relevant is given by $\pi_j$.
Those gates learn collectively which weights are relevant.
The expected objective is then
\begin{align}
    \Ls =\E_{q(\vg\g\pi)}\left[\frac{1}{N}\sum\limits_{i=1}^{N} \norm{\vy_i-\mathrm{iEQL}(\vx_i,\mW)}^2_2\right]
    +\lambda\sum\limits_{u=1}^{U}c_u\sum\limits_{j=1}^{\abs{W_u}}\pi_j^u\,.
\end{align}
$\abs{W_u}$ is the total number of weights that correspond to an atomic unit type ($u$).
Further details on the variational optimization scheme are given in appendix~\ref{app:l0_reg}.

%%%%%%%%%%%%%%%%%%%%%%%%%%%%%%%%%%%%%%%%%%%%%%%%%%%%%%%%%%%%%%%%%%%%%%%%%%%%%%%%
\begin{figure}[hbt!]
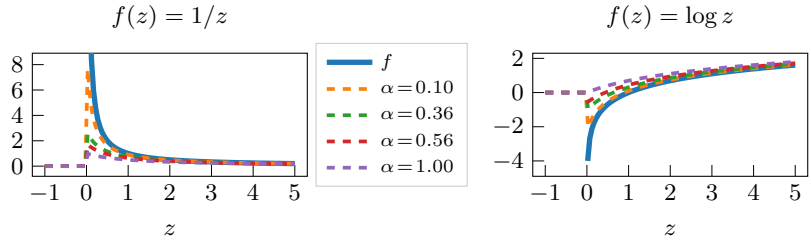

    \begin{small}
    \hfill
    \input{./fig/division}
    \input{./fig/log}
    \hfill
    \end{small}
    \vspace{-.8em}
    \caption{Relaxation of the division (left) and
            logarithm (right)
            for different values of $\alpha$.
 		}
        \label{fig:divergence_relaxation}
\end{figure}
%%%%%%%%%%%%%%%%%%%%%%%%%%%%%%%%%%%%%%%%%%%%%%%%%%%%%%%%%%%%%%%%%%%%%%%%%%%%%%%
\subsection{Atomic Units with Singularities}%\label{sec:singularities}
Our algorithm is designed for atomic units with singularities\footnote{we refer to them as singular units}
$f:\; \mathbb{D} \to \R$
on a half-bounded domain $\mathbb{D}\equiv (a,\infty)$
exhibiting a singularity at $a$
(e.g.\ $\log: (0,\infty) \to \R$ with a singularity at $a=0$).
It also applies to $\mathrm{division}$ units $\nicefrac{c}{d}$
under the assumption that real systems do not diverge in the domain of
application. Then, it is sufficient to only consider the
positive branch of the hyperbola $\nicefrac{1}{d}$ and let $c$ choose the sign.
Cascading transformation of intermediate results in the deep architecture of the \bEQL
can project values outside of the constrained input domains of singular units.
Therefore, the domain $\mathbb{D}$ of a singular unit is continuously extended to $\mathbb{R}$
to avoid forbidden inputs during training (see \eqref{eq:divergences_relaxation}), similar to~\citet{Sahoo2018}.
During training, an additional penalty $\Ls_{\text{su}}$ is necessary
to constrain the solution space of the \bEQL to networks that respect the
domain $\mathbb{D}$ of all its singular units $f_u(z_u)$
\begin{equation}
    \Ls_{\text{su}} =\frac{1}{N} \sum\limits_{i=1}^{N}\sum\limits_{u\in \mathcal{SU}} \Ls_{\text{su}}^{\text{u}}(z_i^u)\,,\qquad
                                   \Ls_{\text{su}}^{\text{u}}(z_i^u) = \begin{cases}
                                                                    0,       &\text{for } z_i^u>a,\\
                                                                    |a-z_i^u|,   &\text{for } z_i^u\leq a\,.
                                                                \end{cases}\label{eq:divergences_penalty}
\end{equation}
The set of all singular units in the \bEQL is given by $\mathcal{SU}$.
Still, training can be corrupted by unconstrained values and exploding gradients in the neighborhood of
the singularities.
Therefore, we propose a learnable relaxation of the singular units
that does not compromise the final symbolic equation.
It is shown in \figref{fig:divergence_relaxation} for
the logarithm and division.
% (see \eqref{eq:divergences_relaxation}).
The input $z$ is shifted \(\hat z = z + \alpha\) by a learnable positive relaxation-parameter
$\alpha=\log(1+e^{\hat\alpha})>0$ %\footnote{to assure positivity we choose $\alpha=\log(1+e^{\hat\alpha})$}
\begin{equation}
\hat f(\hat z) = \begin{cases}
                f(z+\alpha),    & \text{for } z>a,\\
                0,              & \text{for } z\leq a
            \end{cases}\label{eq:divergences_relaxation}\;.
\end{equation}
Since the input $z=Wy+b$ is an affine transformation of the previous layer
with bias $b$ the relaxation-parameter $\alpha$ can be safely added to the bias and does not affect
the structure of the final equation.
This method assures that, during and after training, the maximum absolute
value in the neighborhood of the singularity is smaller than $|f(a+\alpha)|$
and the maximum value of the derivative is smaller than $|\partial
f(a+\alpha)|$ on the training dataset.
The final objective function is thus a sum of data loss $\Ls_{\text{D}}$,
regularization $\Ls_{\text{C}}$
and domain penalty $\Ls_{\text{su}}$ with domain penalty strength $\delta$
\begin{align}
    \Ls =\Ls_{\text{D}}+ \delta \,\Ls_{\text{su}}+\reg\, \Ls_{\text{C}} \,.
\end{align}
In order to guarantee that the domain constraints on singular units
also hold on the extrapolation area and that the order of magnitude does not
change drastically on extrapolation data, intrinsic penalty epochs are
necessary as proposed in \citet{Sahoo2018}.
The bound-penalty $\Ls_{\text{bound}}$ is given by
\begin{align}
    \Ls_{\text{bound}} = \frac{1}{N}\sum\limits_{i=1}^N \argmax(\abs{\mathrm{iEQL}(\vx_i, \mW)}-B,0)
\end{align}
For an intrinsic penalty epoch, we randomly sample $N$ input data points from
the test area without labels. Then the \bEQL is trained on the loss function
\begin{align}
    \Ls_{\text{penalty}}=\delta\,\Ls_{\text{su}}+ \delta \,\Ls_{\text{bound}}\,.
\end{align}

%%%%%%%%%%%%%%%%%%%%%%%%%%%%%%%%%%%%%%%%%%%%%%%%%%%%%%%%%%%%%%%%%%%%%%%%%%%%%%%%%%%%%%%%%%%%%%%%%%%%%%%%

\section{Experiments}\label{sec:experiments}

We demonstrate the application of the \bEQL\ to four
different use cases.
The first one is to learn complex equations on simulated data of several equations.
The second use case is a simulated ambiguity dataset, with which we
demonstrate how to incorporate expert knowledge to influence the outcome of the
equation learner. Additionally, we study the relative frequency of selected atomic units
in found equations with and without expert knowledge on a simulated dataset and a real world dataset.
Use cases three and four both address real-world applications in industry
determining expressions for the power loss of an electric machine and a torque
model of a combustion engine.
% \matthias{is it sufficient to mention the architecture and training procedure just in the appendix?}
All experiments are performed using an \bEQL with four hidden layers.
Each hidden layer has
\(\{\cos,\,\exp{\;},\,\log,\, \sqrt{\phantom{x}},\,x^2,\, *,\,/\}\)
as atomic units, and each atomic unit is applied four times in each layer.
Combinations from \eqref{eq:prohibited_components} are prohibited.
We compare the \bEQL to five different algorithms:
\begin{itemize}
    \item \dEQL, a state-of-the-art method from \citet{Sahoo2018} with atomic unit types
$\{\sin, \cos, *, \text{identity}\}$ in each hidden layer and division in the final layer,
    \item a multi-layer perceptron (MLP) with $\tanh$ activation functions and five hidden layers
with 50 neurons each,
    \item a genetic algorithm (\textsc{PySR},\cite{pysr}) with two different configurations \pysrA and \pysrB,
    \item a Gaussian Process (GP) for the real world datasets, calculated with \textsc{ascmo} \citep{Ascmo2015}, which is a standard tool from the engineering domain,
    \item the mean predictor (MP) on the train set.
\end{itemize}
Further details on training and parameter settings are outlined in appendix~\ref{app:ieql_training}.
All experiments are executed five times, and we report median, minimum and maximum (in sub and superscript)
of root mean squared error on test datasets.

\subsection{Learning Complex Equations}\label{sec:complex_eq}
We evaluate the \bEQL on two different sets of ground truth equations shown in \tableref{ta:datasets}:
(S0--S6) with $\{\pm,\sin,\cos,*,/\}$ as operations and (A1--A4) also including $\{\log,\exp\}$.
A two dimensional contour plot of all datasets is shown in appendix \figref{fig:dataset_contour}.
Most functions are gathered from different papers on symbolic regression.
A star ($^*$) indicates that input and output values were scaled.
We closely follow the evaluation strategy proposed in \citet{Sahoo2018} with their model selection
criteria \ValIE with 40 extrapolation points (see \secref{sec:model_selection}) on the found equations
of \bEQL, \dEQL and \pysr.
Training datasets consist of $10^4$ randomly sampled data points in the train domain.
Outputs are corrupted with standard normal noise of standard deviation $\gamma=0.01$.
For validation $10\%$ of the training dataset is used.
Test datasets consist of $10^3$ randomly sampled data points from the test domain (see \tableref{ta:datasets}).
Note that for the equations S0--S6 and A1 with their four dimensional input, the extrapolation area of the test dataset is 15 times larger than its interpolation area.
The included extrapolation domain is a good indicator for whether or not
the ground truth has been found.
%%%%%%%%%%%%%%%%%%%%%%%%%%%%%%%%%%%%%%%%%%%%%%%%%%%%%%%%%%%%%%%%%%%%%%%%%%%%%%%%
\begin{table}
   \centering
    \caption{Complex equations on which the \bEQL is evaluated. Most functions are
    gathered from different papers on symbolic regression. A star ($^*$) indicates that input and
    output were scaled.
     }
    \label{ta:datasets}
    \begin{adjustbox}{max width=\linewidth}
            {\tabulinesep=1.1mm
\begin{tabu}{@{\ }ll@{}c@{}c@{\quad}c@{\ }}
    \toprule
    Data- & {Ground truth}               & motivated               & \multicolumn{2}{c}{Domain}   \\
    set   &  equation                   & from                     & train & test\\
    \midrule
    S0      &  $y = (1-x_2^2)/(\sin(2\pi\,x_1) + 1.5)$
        &           & $[-1,1]^4$     & $[-2,2]^4$   \\
    S1      &  $y = \left[\sin(\pi x_1) + \sin(2\pi x_2 + \nicefrac{\pi}{8})
                   +x_2 -x_3x_4\right]/3 $
       & \citet{Sahoo2018}        & $[-1,1]^4$     & $[-2,2]^4$   \\
    S2      &  $y = \left[\sin(\pi x_1)\!+\! x_2\! \cos(2\!\pi x_1 +
                   \nicefrac{\pi}{4})\!+\!x_3\!-\!x_4^2\right]/3 $
        & \citet{Sahoo2018}          & $[-1,1]^4$     & $[-2,2]^4$   \\
    S3      &  $y = \left[(1+x_2) \sin(\pi x_1) + x_2x_3x_4\right]/3$
        & \citet{Sahoo2018}          & $[-1,1]^4$     & $[-2,2]^4$   \\
    S4$^*$      &  $y = (3.0375\, x_1 x_2 + 5.5 \sin{(\nicefrac{9}{4}\,(x_1-\nicefrac{2}{3})(x_2-\nicefrac{2}{3}))})/5$
        &  \citet{Jin2019}    & $[-1,1]^4$     & $[-2,2]^4$   \\
    S5$^*$      &  $y = \frac{(5x_1)^4}{(5x_1)^4+1}+\frac{(5x_2)^4}{(5x_2)^4+1}$
        & \citet{Trujillo2016}         & $[-1,1]^4$     & $[-2,2]^4$   \\
    \multirow{2}{*}{S6$^*$}  &$y =  ((1-x_1)^2 + (1-x_3)^2 + 100(x_2-x_1^2)^2$
        &\multirow{2}{*}{4D-Rosenbrock}   &\multirow{2}{*}{ $[-1,1]^4$ }&\multirow{2}{*}{$[-2,2]^4$}\\
                           &$\quad\quad+ 100(x_4-x_3^2)^2)/1500$&\\
    % &  $y =  ((1-x_1)^2 + (1-x_3)^2 + 100(x_2-x_1^2)^2$&
        % &           &\multirow{2}{*}{ $[-1,1]^4$ }    &\multirow{}{*}{ $[-2,2]^4$}   \\
                     % $\qqaud+ 100(x_4-x_3^2)^2)/1500$ \\
    \midrule
    A1$^*$      &  $y = (1.5 \e^{1.5x_1}+ 5 \cos{(3\, x_2)})/10$
        &   \citet{Jin2019}   & $[-1,1]^4$     & $[-2,2]^4$   \\
    A2$^*$      &  $y = \log(2 x_2+1)-\log(4 x_1^2 +1)$
        &  \citet{Trujillo2016}         & $[0,1]^2$     & $[0,2]^2$   \\
    A3$^*$      &  $y =\frac{ \exp(- (4x_1-0.4)^2)}{1.2+(4 x_2-1.9)^2}$
        &   \citet{Trujillo2016}         & $[0,1]^2$     & $[0,2]^2$   \\
    A4$^*$      &  $y = \frac{x_3+1}{3\pi^2}\frac{(5x_1+1)^3}{\exp\left(\frac{(x_3+1)(5x_1+1)}
                    {(x_4+1)(0.5x_2+1)}\right)-1}$
        &    {\scriptsize Udrescu and Tegmark (2020)}     & $[0,1]^4$     & $[0,2]^4$   \\ %I.41.16
    \bottomrule
\end{tabu}}

    \end{adjustbox}
\end{table}
%%%%%%%%%%%%%%%%%%%%%%%%%%%%%%%%%%%%%%%%%%%%%%%%%%%%%%%%%%%%%%%%%%%%%%%%%%%%%%%%
%
%
%
\begin{table*}
  \centering
    \caption{Learning complex equations S0-S6: median, minimum and maximum (in sub and superscript)
    of root mean squared error (RMSE) on test datasets.
      Baselines are the \dEQL from \citet{Sahoo2018}, a genetic algorithm (GA) from \citet{pysr},
      a multi-layer perceptron (MLP) and the mean predictor (MP) on the train dataset.}
    \label{ta:S_5r_VE}
    \begin{adjustbox}{max width=\linewidth}
        % \begin{tabular}{@{\ }l *{7}{l}@{\ }}
\begin{tabular}{l
        r
  @{\extracolsep{\fill}}
    l
        r
  @{\extracolsep{\fill}}
    l
        r
  @{\extracolsep{\fill}}
    l
        r
  @{\extracolsep{\fill}}
    l
        r
  @{\extracolsep{\fill}}
    l
        r
  @{\extracolsep{\fill}}
    l
        r
  @{\extracolsep{\fill}}
    l
    }
\toprule
{} & \multicolumn{2}{c}{S0} & \multicolumn{2}{c}{S1} & \multicolumn{2}{c}{S2} & \multicolumn{2}{c}{S3} & \multicolumn{2}{c}{S4} & \multicolumn{2}{c}{S5} & \multicolumn{2}{c}{S6} \\
% {} &                      VE &                      VE &                      VE &                      VE &                      VE &                      VE &                      VE \\
\midrule
MP    &               1.57& &                  0.66& &                    0.75& &       0.64 &         &           1.17 &  &                  0.44& &                    0.82& \\
MLP   & 1.11 &\mm{1.04}{1.17} & 0.45 & \mm{0.35}{0.54} & 0.41 &\mm{0.39}{0.42} & 0.36 &\mm{0.34}{0.37} &  0.65&\mm{0.57}{0.70} & 0.07& \mm{0.02}{0.15} &  0.65&\mm{0.65}{0.66} \\
\pysrA & 1.07 &\mm{0.91}{1.58} & 0.77 & \mm{0.77}{0.77} & 0.33 &\mm{0.01}{0.35} & 0.01 &\mm{0.01}{0.47} &  1.20&\mm{1.07}{1.23} & 0.29& \mm{0.29}{0.29} &  0.82&\mm{0.82}{0.82} \\
\pysrB & 0.39 &\mm{0.01}{1.47} & 0.66& \mm{0.01}{0.77} & 0.37& \mm{0.01}{0.77} & 0.60& \mm{0.58}{1.78} & 0.01& \mm{0.01}{0.01} & 0.08& \mm{0.01}{0.42} &  4.26& \mm{0.74}{\text{inf}} \\
\dEQL & 0.01 &\mm{0.01}{0.12} & 0.01 & \mm{0.01}{0.17} & 0.01 &\mm{0.01}{0.05} & 0.01 &\mm{0.01}{0.01} &  0.01&\mm{0.01}{1.38} & 0.01& \mm{0.01}{0.01} &  0.01&\mm{0.01}{0.02} \\
\bEQL & 0.01 &\mm{0.01}{0.01} & 0.01 & \mm{0.01}{0.01} & 0.01 &\mm{0.01}{0.01} & 0.01 &\mm{0.01}{0.01} &  0.01&\mm{0.01}{0.01} & 0.01& \mm{0.01}{0.01} &  0.01&\mm{0.01}{0.01} \\
\bottomrule
\end{tabular}

    \end{adjustbox}
\end{table*}

All three network-based models (MLP, \dEQL, \bEQL) can fit the training data perfectly.
Their prediction error on training and validation data is at noise level for all datasets S0--S6, and A1--A4.
% The genetic algorithm (\pysrA) is only for the datasets S2, S3, A1, and A2 at noise level on the validation data.
\Tableref{ta:S_5r_VE} shows the results on the test datasets for S0--S6.
Unsurprisingly, the MLP is not able to capture the data-generating function (ground truth) by any means.
Therefore, it is not able to perform well on the test domain, which includes large extrapolation domains,
except for equation A3, which is mainly zero in the extrapolation domain.
In contrast, \pysr, \dEQL, and \bEQL can learn meaningful equations, which
can extrapolate on the test domain.
If the correct equation has been identified by the \ValIE criteria,
the test RMSE is expected to be at noise level.
The genetic algorithm (\pysrA) captures the data generating function on the datasets S2 and S3 at least in one experiment
and \pysrB on the datasets S0, S1, S2, S4 and S5.
On datasets S0--S6 the \bEQL captures the data generating function reliably for all five runs as shown in \tableref{ta:S_5r_VE}.
Despite its larger expressivity, with
\(\{\cos,\,\exp{\;},\,\log,\, \sqrt{\phantom{x}},\,x^2,\, *,\,/\}\) atomic units,
it even outperforms the \dEQL architecture, which uses just the necessary atomic units
\(\{\cos,\,\sin,\, *,\,/\}\).
Equations A1--A4 are more difficult to learn.
\bEQL outperforms all baselines  as shown in \tableref{ta:A_5r_VE}.
\begin{table}
   \centering
    \caption{Learning complex equations (A1--A4): median, minimum and maximum (in sub and superscript)
    of root mean squared error (RMSE) on test datasets.
      Baselines are the \dEQL from \citet{Sahoo2018}, a geneatic algorithm (GA) from \citet{pysr}, a multi-layer perceptron (MLP) and the mean predictor (MP) on the train dataset.}
    \label{ta:A_5r_VE}
    \begin{adjustbox}{max width=\linewidth}
            \begin{tabular}{l
        r
  @{\extracolsep{\fill}}
    l
        r
  @{\extracolsep{\fill}}
    l
        r
  @{\extracolsep{\fill}}
    l
        r
  @{\extracolsep{\fill}}
    l
    }
\toprule
    {} &                      \multicolumn{2}{c}{A1} &                      \multicolumn{2}{c}{A2} &                      \multicolumn{2}{c}{A3} &                      \multicolumn{2}{c}{A4} \\
% {} &                      VE &                      VE &                      VE &                      VE \\
\midrule
MP    &        0.88& &            1.09& &                0.19 &&                 0.66 &\\
MLP   & 0.67 &\mm{0.64}{0.70} & 0.24 &\mm{0.22}{0.25} & 0.01 &\mm{0.01}{0.02} &  0.39&\mm{0.37}{0.40} \\
\pysrA & 0.82 &\mm{0.28}{1.11} & 0.35 &\mm{0.26}{0.62} & 0.14 &\mm{0.14}{0.19} &  0.60&\mm{0.56}{0.66} \\
\pysrB & 0.28& \mm{0.04}{0.45} & 0.10& \mm{0.06}{0.51} & 0.10& \mm{0.04}{0.77} &0.35&  \mm{0.28}{0.60} \\
\dEQL & 0.09 &\mm{0.09}{0.11} & 0.09 &\mm{0.06}{0.19} & 0.02 &\mm{0.01}{0.33} &  0.36&\mm{0.22}{1.89} \\
\bEQL & 0.02 &\mm{0.01}{0.06} & 0.03 &\mm{0.02}{0.05} & 0.01 &\mm{0.01}{0.02} &  0.07&\mm{0.07}{0.13} \\
\bottomrule
\end{tabular}

        \end{adjustbox}
    \end{table}
    %%%%%%%%%%%%%%%%%%%%%%%%%%%%%%%%%%%%%%%%%%%%%%%%%%%%%%%%%%%%%%%%%%%%%%%%%%%%%%%%
    %%%%%%%%%%%%%%%%%%%%%%%%%%%%%%%%%%%%%%%%%%%%%%%%%%%%%%%%%%%%%%%%%%%%%%%%%%%%%%%%
%
%
%
    \subsection{Expert Knowledge}
    \label{sec:ambiguity}
    \begin{figure}
            \centering
            \subfigure[relative frequency of atomic units]{
                % This file was created by tikzplotlib v0.9.1.
\begin{tikzpicture}

\definecolor{color0}{rgb}{0.890196078431372,0.466666666666667,0.76078431372549}
\definecolor{color1}{rgb}{0.172549019607843,0.627450980392157,0.172549019607843}

\begin{axis}[
height=0.5\figheight,
legend cell align={left},
legend style={fill opacity=0.9, draw opacity=1, text opacity=1, draw=white!80!black},
tick align=outside,
tick pos=left,
width=\figwidth,
x grid style={white!69.0196078431373!black},
xlabel={ ratio \(\displaystyle r=c_{\cos}/c_{x^2}\)},
xmin=-1.175, xmax=19.175,
xtick style={color=black},
xtick={0,1,2,3,4,5,6,7,8,9,10,11,12,13,14,15,16,17,18},
xticklabel style = {rotate=80.0},
xticklabels={\(\displaystyle 0.10\),\(\displaystyle 0.11\),\(\displaystyle 0.12\),\(\displaystyle 0.14\),\(\displaystyle 0.17\),\(\displaystyle 0.20\),\(\displaystyle 0.25\),\(\displaystyle 0.33\),\(\displaystyle 0.50\),\(\displaystyle 1.00\),\(\displaystyle 2.00\),\(\displaystyle 3.00\),\(\displaystyle 4.00\),\(\displaystyle 5.00\),\(\displaystyle 6.00\),\(\displaystyle 7.00\),\(\displaystyle 8.00\),\(\displaystyle 9.00\),\(\displaystyle 10.00\)},
y grid style={white!69.0196078431373!black},
ylabel={[\%]},
ymin=0, ymax=105,
ytick style={color=black},
ytick={0,20,40,60,80,100,120},
yticklabels={\(\displaystyle 0\),\(\displaystyle 20\),\(\displaystyle 40\),\(\displaystyle 60\),\(\displaystyle 80\),\(\displaystyle 100\),\(\displaystyle 120\)},
]
\draw[draw=none,fill=color1,postaction={pattern=north east lines}] (axis cs:-0.25,66.7635504592026) rectangle (axis cs:0.25,97.4783228044097);
\addlegendimage{ybar,ybar legend,draw=none,fill=color1,fill opacity=0.6,postaction={pattern=north east lines}};
\addlegendentry{\(\displaystyle x^2\)}
\draw[draw=none,fill=color1,postaction={pattern=north east lines}] (axis cs:0.75,69.9119570905285) rectangle (axis cs:1.25,98.3772457701029);
\draw[draw=none,fill=color1,postaction={pattern=north east lines}] (axis cs:1.75,66.5718635939224) rectangle (axis cs:2.25,98.7575333163569);
\draw[draw=none,fill=color1,postaction={pattern=north east lines}] (axis cs:2.75,71.4307995925643) rectangle (axis cs:3.25,98.0724789915966);
\draw[draw=none,fill=color1,postaction={pattern=north east lines}] (axis cs:3.75,68.4710884353741) rectangle (axis cs:4.25,98.6235827664399);
\draw[draw=none,fill=color1,postaction={pattern=north east lines}] (axis cs:4.75,67.5730911600477) rectangle (axis cs:5.25,97.6224459920112);
\draw[draw=none,fill=color1,postaction={pattern=north east lines}] (axis cs:5.75,57.4094001236858) rectangle (axis cs:6.25,98.510101010101);
\draw[draw=none,fill=color1,postaction={pattern=north east lines}] (axis cs:6.75,55.5675833287774) rectangle (axis cs:7.25,98.5590818799774);
\draw[draw=none,fill=color1,postaction={pattern=north east lines}] (axis cs:7.75,51.5055674010898) rectangle (axis cs:8.25,98.0164652925847);
\draw[draw=none,fill=color1,postaction={pattern=north east lines}] (axis cs:8.75,36.2820512820513) rectangle (axis cs:9.25,98.9871794871795);
\draw[draw=none,fill=color1,postaction={pattern=north east lines}] (axis cs:9.75,21.371487989135) rectangle (axis cs:10.25,99.6428571428571);
\draw[draw=none,fill=color1,postaction={pattern=north east lines}] (axis cs:10.75,12.4102431565118) rectangle (axis cs:11.25,100);
\draw[draw=none,fill=color1,postaction={pattern=north east lines}] (axis cs:11.75,14.333436404865) rectangle (axis cs:12.25,99.5685425685426);
\draw[draw=none,fill=color1,postaction={pattern=north east lines}] (axis cs:12.75,18.5997134910178) rectangle (axis cs:13.25,99.5915678524374);
\draw[draw=none,fill=color1,postaction={pattern=north east lines}] (axis cs:13.75,15.1949418740464) rectangle (axis cs:14.25,100);
\draw[draw=none,fill=color1,postaction={pattern=north east lines}] (axis cs:14.75,13.5323418411654) rectangle (axis cs:15.25,100);
\draw[draw=none,fill=color1,postaction={pattern=north east lines}] (axis cs:15.75,14.7382420320362) rectangle (axis cs:16.25,99.734878240377);
\draw[draw=none,fill=color1,postaction={pattern=north east lines}] (axis cs:16.75,14.7583503098209) rectangle (axis cs:17.25,99.7315592903828);
\draw[draw=none,fill=color1,postaction={pattern=north east lines}] (axis cs:17.75,15.4974927033751) rectangle (axis cs:18.25,99.8663101604278);
\draw[draw=none,fill=color0,postaction={pattern=crosshatch dots}] (axis cs:-0.25,0) rectangle (axis cs:0.25,66.7635504592026);
\addlegendimage{ybar,ybar legend,draw=none,fill=color0,fill opacity=0.6,postaction={pattern=crosshatch dots}};
\addlegendentry{\(\displaystyle \cos\)}
\draw[draw=none,fill=color0,postaction={pattern=crosshatch dots}] (axis cs:0.75,0) rectangle (axis cs:1.25,69.9119570905285);
\draw[draw=none,fill=color0,postaction={pattern=crosshatch dots}] (axis cs:1.75,0) rectangle (axis cs:2.25,66.5718635939224);
\draw[draw=none,fill=color0,postaction={pattern=crosshatch dots}] (axis cs:2.75,0) rectangle (axis cs:3.25,71.4307995925643);
\draw[draw=none,fill=color0,postaction={pattern=crosshatch dots}] (axis cs:3.75,0) rectangle (axis cs:4.25,68.4710884353741);
\draw[draw=none,fill=color0,postaction={pattern=crosshatch dots}] (axis cs:4.75,0) rectangle (axis cs:5.25,67.5730911600477);
\draw[draw=none,fill=color0,postaction={pattern=crosshatch dots}] (axis cs:5.75,0) rectangle (axis cs:6.25,57.4094001236858);
\draw[draw=none,fill=color0,postaction={pattern=crosshatch dots}] (axis cs:6.75,0) rectangle (axis cs:7.25,55.5675833287774);
\draw[draw=none,fill=color0,postaction={pattern=crosshatch dots}] (axis cs:7.75,0) rectangle (axis cs:8.25,51.5055674010898);
\draw[draw=none,fill=color0,postaction={pattern=crosshatch dots}] (axis cs:8.75,0) rectangle (axis cs:9.25,36.2820512820513);
\draw[draw=none,fill=color0,postaction={pattern=crosshatch dots}] (axis cs:9.75,0) rectangle (axis cs:10.25,21.371487989135);
\draw[draw=none,fill=color0,postaction={pattern=crosshatch dots}] (axis cs:10.75,0) rectangle (axis cs:11.25,12.4102431565118);
\draw[draw=none,fill=color0,postaction={pattern=crosshatch dots}] (axis cs:11.75,0) rectangle (axis cs:12.25,14.333436404865);
\draw[draw=none,fill=color0,postaction={pattern=crosshatch dots}] (axis cs:12.75,0) rectangle (axis cs:13.25,18.5997134910178);
\draw[draw=none,fill=color0,postaction={pattern=crosshatch dots}] (axis cs:13.75,0) rectangle (axis cs:14.25,15.1949418740464);
\draw[draw=none,fill=color0,postaction={pattern=crosshatch dots}] (axis cs:14.75,0) rectangle (axis cs:15.25,13.5323418411654);
\draw[draw=none,fill=color0,postaction={pattern=crosshatch dots}] (axis cs:15.75,0) rectangle (axis cs:16.25,14.7382420320362);
\draw[draw=none,fill=color0,postaction={pattern=crosshatch dots}] (axis cs:16.75,0) rectangle (axis cs:17.25,14.7583503098209);
\draw[draw=none,fill=color0,postaction={pattern=crosshatch dots}] (axis cs:17.75,0) rectangle (axis cs:18.25,15.4974927033751);
\end{axis}

\end{tikzpicture}
                                     \label{fig:amb_stats}}\\
            \subfigure[\ValISparse with complexity factors]{
                % This file was created by tikzplotlib v0.9.1.
\begin{tikzpicture}

\definecolor{color0}{rgb}{0.890196078431372,0.466666666666667,0.76078431372549}
\definecolor{color1}{rgb}{0.172549019607843,0.627450980392157,0.172549019607843}

\begin{axis}[
height=0.5\figheight,
legend cell align={left},
legend style={fill opacity=0.5, draw opacity=1, text opacity=1, draw=white!80!black,/tikz/column 2/.style={column sep=5pt}, font=\scriptsize},
log basis x={10},
tick pos=left,
width=0.45\figwidth,
x grid style={white!50.1960784313725!black},
xlabel={ ratio \(\displaystyle r=c_{\cos}/c_{x^2}\)},
xmin=0.0794328234724281, xmax=12.5892541179417,
xmode=log,
xtick style={color=black},
xticklabels={\(\displaystyle 0.1\),\(\displaystyle 0.3\),\(\displaystyle 1.0\),\(\displaystyle 3.0\),\(\displaystyle 10\)},
xtick={0.1,0.3,1,3,10},
y grid style={white!50.1960784313725!black},
ylabel={\(\displaystyle \#\) units},
ymin=-0.2, ymax=2.5,
ytick style={color=black},
]
\addplot [line width=1.4pt, color0, dotted, mark=*, mark size=1, mark options={solid}]
table {%
0.1 1
0.111111111111111 1
0.125 1
0.142857142857143 1
0.166666666666667 1
0.2 1
0.25 1
0.333333333333333 1
0.5 1
1 0
2 0
3 0
4 0
5 0
6 0
7 0
8 0
9 0
10 0
};
\addlegendentry{$\cos$}
\addplot [line width=1.4pt, color1, dashed, mark=*, mark size=1, mark options={solid}]
table {%
0.1 0
0.111111111111111 0
0.125 0
0.142857142857143 0
0.166666666666667 0
0.2 0
0.25 0
0.333333333333333 0
0.5 0
1 1
2 1
3 1
4 1
5 1
6 1
7 1
8 1
9 1
10 1
};
\addlegendentry{$x^2$}
\end{axis}

\end{tikzpicture}
                                     \label{fig:amb_VC}}\hfill
            \subfigure[\ValISparse without complexity factors]{
                % This file was created by tikzplotlib v0.9.1.
\begin{tikzpicture}

\definecolor{color0}{rgb}{0.890196078431372,0.466666666666667,0.76078431372549}
\definecolor{color1}{rgb}{0.172549019607843,0.627450980392157,0.172549019607843}

\begin{axis}[
height=0.5\figheight,
legend cell align={left},
legend style={fill opacity=0.5, draw opacity=1, text opacity=1, draw=white!80!black,/tikz/column 2/.style={column sep=5pt}, font=\scriptsize},
log basis x={10},
tick pos=left,
width=0.45\figwidth,
x grid style={white!50.1960784313725!black},
xlabel={ ratio \(\displaystyle r=c_{\cos}/c_{x^2}\)},
xmin=0.0794328234724281, xmax=12.5892541179417,
xmode=log,
xtick style={color=black},
xticklabels={\(\displaystyle 0.1\),\(\displaystyle 0.3\),\(\displaystyle 1.0\),\(\displaystyle 3.0\),\(\displaystyle 10\)},
xtick={0.1,0.3,1,3,10},
y grid style={white!50.1960784313725!black},
ylabel={\(\displaystyle \#\) units},
ymin=-0.2, ymax=2.5,
ytick style={color=black},
]
\addplot [line width=1.4pt, color0, dotted, mark=*, mark size=1, mark options={solid}]
table {%
0.1 0
0.111111111111111 1
0.125 0
0.142857142857143 0
0.166666666666667 1
0.2 0
0.25 0
0.333333333333333 0
0.5 0
1 0
2 0
3 0
4 0
5 0
6 0
7 0
8 0
9 0
10 0
};
\addlegendentry{$\cos$}
\addplot [line width=1.4pt, color1, dashed, mark=*, mark size=1, mark options={solid}]
table {%
0.1 1
0.111111111111111 0
0.125 1
0.142857142857143 1
0.166666666666667 0
0.2 1
0.25 1
0.333333333333333 1
0.5 1
1 1
2 1
3 1
4 1
5 1
6 1
7 1
8 1
9 1
10 1
};
\addlegendentry{$x^2$}
\end{axis}

\end{tikzpicture}
                                     \label{fig:amb_VC_cf}}
        \caption{%
        The effect of expert knowledge
         is quantified by means of an ambiguous dataset with its
        ground truth given by \eqref{eq:ambiguity}.
        For each complexity cost ratio $r=\nicefrac{c_{\cos}}{c_{x^2}}$
        the relative frequency of atomic units is calculated for each found equation.
        Panel (a) shows the average of all those relative frequencies for $\cos$ and $x^2$ units.
        The number of $\cos$ and $x^2$ units appearing in the selected equation
        depending on the complexity cost ratio is shown in panel (b)
        selected with \ValISparse with complexity factors and
        in panel (c) with \ValISparse without complexity factors.
        }
        \label{fig:ambiguity}
    \end{figure}
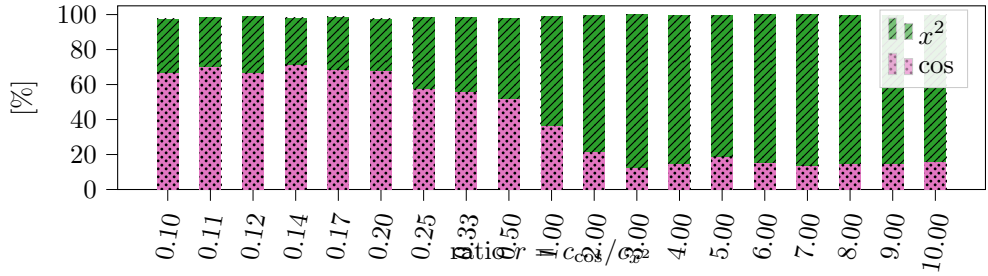
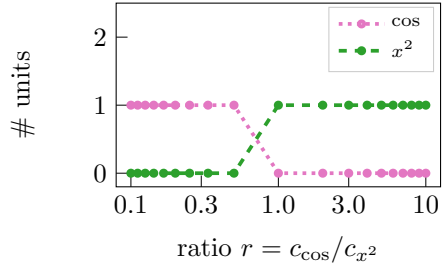
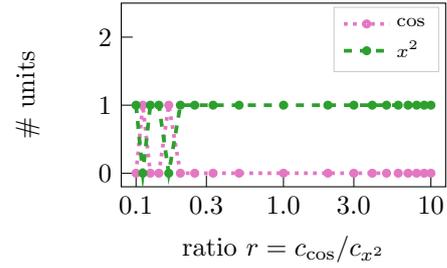
    Expert knowledge is especially important if there are several possible
    equations for an ambiguous dataset.
    Incorporating this knowledge into the complexity measure as well as the
    regularization can decide whether the right equation is found.
    The following experiment studies how different sets of complexity factors affect
    the relative frequency of atomic units in the set of found equations as well as the
    importance of complexity factors for the selection criteria \ValISparse, which is
    applied to the set of found equations.
    The selection criteria \ValIE is not suited for this experiment,
    since it ignores the model complexity.
    In order to construct an ambiguous dataset we choose the ground truth
    \begin{align}\label{eq:ambiguity}
        y &= 8\cos(0.5\, x)-4%+ \epsilon\,,\q
%        \epsilon \sim \gN(0, 0.01^2)\,
    \end{align}
    which, close to $x=0$, can be modeled equivalently by a $\cos$ function
    or a polynomial of degree 2.
    The output is corrupted with standard normal noise $\gN(0,0.01^2)$.
    The training dataset consists of $10^4$ randomly sampled data points on the train domain $[-1,1]$.
    It resembles a parabola.
    We study 9 configurations that prefer a periodic structure $c_{\cos}\in\{\nicefrac{1}{2},\nicefrac{1}{3},\dots,\nicefrac{1}{10}\}$ and
    9 configurations that prefer a polynomial structure $c_{x^2}\in\{\nicefrac{1}{2},\nicefrac{1}{3},\dots,\nicefrac{1}{10}\}$ as well as
    the plain configuration.
    For each complexity cost ratio $r=\nicefrac{c_{\cos}}{c_{x^2}}$ we calculate the relative frequency of atomic units
    for each of the 78 found equations (number of different regularization strength, see appendix~\ref{app:ieql_training}). \Figref{fig:amb_stats} shows the average of all those
    relative frequencies for $\cos$ and $x^2$ units.
    As desired the relative frequencies follow the prior.
    Ratios with $r<1$ prefer  $\cos$ units and
    ratios with $r>1$ prefer  $x^2$ units.
    Without any further information ($r=1$) a polynomial solution seems very plausible.
    It turns out that \bEQL does not converge to all kinds of simple ambiguous solutions
    with the same probability. Often, certain solutions are more likely to converge to.
    In this experiment, for instance, \bEQL predominantly uses $x^2$ units without a biased preference ($r=1$).
    \Figref{fig:amb_VC} shows the number of $\cos$ and $x^2$ units appearing in the selected equation
    depending on the complexity cost ratio.
    Ratios with $r<1$ prefer a periodic equation and
    ratios with $r>1$ prefer a polynomial equation.
    As desired, the prior is followed in case of such a strong ambiguity.
    % Under the precondition that expert knowledge prefers a periodic equation
     % the \ValISparse criteria identifies such periodic equation reliably.
    Without complexity factors a polynomial solution is mostly selected 
    as shown in \figref{fig:amb_VC_cf}.
    Hence, complexity factors are crucial to select the correct equation with respect to the prior.
    Since in this case both types of found equations have the same complexity we found that the \textsc{Adam} optimizer
    converges better for $x^2$ atomic units. This  leads to the preference of polynomial solutions.

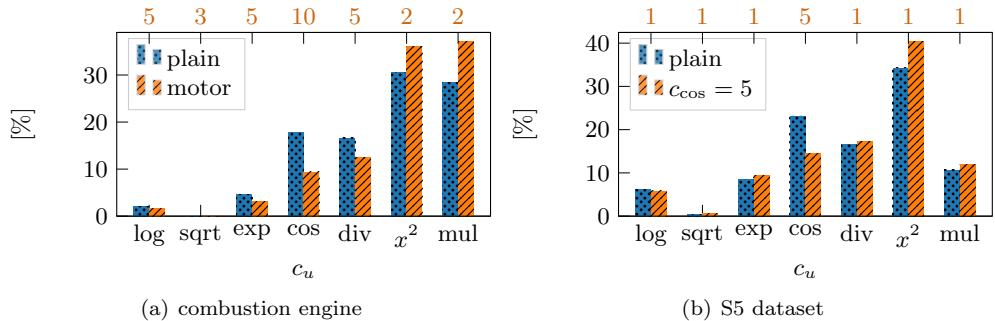
\begin{figure}[t]
    \begin{small}
   \centering
   \subfigure[combustion engine]{\small% This file was created by tikzplotlib v0.9.1.
\begin{tikzpicture}

\definecolor{color0}{rgb}{0.12156862745098,0.466666666666667,0.705882352941177}
\definecolor{color1}{rgb}{1,0.498039215686275,0.0549019607843137}
\definecolor{color2}{rgb}{0.8, 0.39843137254902006, 0.04392156862745096}
 %   \begin{axis}[
 %     xtick={0,1,2,3,4,5,6},
 %     xticklabels={log,sqrt,exp,cos,div,\(\displaystyle x^2\),mul},
 %     scale only axis,
 %     xmin=0,xmax=6,
 %     domain=0:6,
 %     axis y line*=right,
 %     axis x line*=top]
 %     % \addplot[red] {x};
 %   \end{axis}

\begin{axis}[
height=0.5\figheight,
width=0.5\figwidth,
legend cell align={left},
legend style={fill opacity=0.8, draw opacity=1, text opacity=1, at={(0.03,0.97)}, anchor=north west, draw=white!80!black},
tick pos=left,
x grid style={white!50.1960784313725!black},
xmin=-0.63, xmax=6.63,
xtick style={color=black},
xtick={0,1,2,3,4,5,6},
xticklabels={log,sqrt,exp,cos,div,\(\displaystyle x^2\),mul},
y grid style={white!50.1960784313725!black},
ylabel={[\%]},
ymin=0, ymax=39.043167675369,
ytick style={color=black},
ytick={0,10,20,30,40},
yticklabels={\(\displaystyle 0\),\(\displaystyle 10\),\(\displaystyle 20\),\(\displaystyle 30\),\(\displaystyle 40\)},
]
\draw[draw=none,fill=color0,postaction={pattern=crosshatch dots}] (axis cs:-0.3,0) rectangle (axis cs:0,2.13650777671267);
\addlegendimage{ybar,ybar legend,draw=none,fill=color0,postaction={pattern=crosshatch dots}};
\addlegendentry{plain}

\draw[draw=none,fill=color0,postaction={pattern=crosshatch dots}] (axis cs:0.7,0) rectangle (axis cs:1,0.0320151746324281);
\draw[draw=none,fill=color0,postaction={pattern=crosshatch dots}] (axis cs:1.7,0) rectangle (axis cs:2,4.54629229894294);
\draw[draw=none,fill=color0,postaction={pattern=crosshatch dots}] (axis cs:2.7,0) rectangle (axis cs:3,17.7092836688245);
\draw[draw=none,fill=color0,postaction={pattern=crosshatch dots}] (axis cs:3.7,0) rectangle (axis cs:4,16.6244644275616);
\draw[draw=none,fill=color0,postaction={pattern=crosshatch dots}] (axis cs:4.7,0) rectangle (axis cs:5,30.4781592602971);
\draw[draw=none,fill=color0,postaction={pattern=crosshatch dots}] (axis cs:5.7,0) rectangle (axis cs:6,28.4732773930288);
\draw[draw=none,fill=color1,postaction={pattern=north east lines}] (axis cs:2.77555756156289e-17,0) rectangle (axis cs:0.3,1.70067531738352);
\addlegendimage{ybar,ybar legend,draw=none,fill=color1,postaction={pattern=north east lines}};
\addlegendentry{motor}

\draw[draw=none,fill=color1,postaction={pattern=north east lines}] (axis cs:1,0) rectangle (axis cs:1.3,0.0266662523678678);
\draw[draw=none,fill=color1,postaction={pattern=north east lines}] (axis cs:2,0) rectangle (axis cs:2.3,3.21520285623465);
\draw[draw=none,fill=color1,postaction={pattern=north east lines}] (axis cs:3,0) rectangle (axis cs:3.3,9.39471080154901);
\draw[draw=none,fill=color1,postaction={pattern=north east lines}] (axis cs:4,0) rectangle (axis cs:4.3,12.4369976482893);
\draw[draw=none,fill=color1,postaction={pattern=north east lines}] (axis cs:5,0) rectangle (axis cs:5.3,36.0417779095385);
\draw[draw=none,fill=color1,postaction={pattern=north east lines}] (axis cs:6,0) rectangle (axis cs:6.3,37.1839692146371);
\end{axis}

\begin{axis}[
% axis x line=top,
axis y line=none,
height=0.5\figheight,
width=0.5\figwidth,
% log basis x={10},
tick pos=right,
x grid style={white!50.1960784313725!black},
% xmin=0.0616595001861482, xmax=16.2181009735893,
% xmode=log,
xlabel={$c_u$},
xmin=-0.63, xmax=6.63,
xtick style={color=black},
xtick={0,1,2,3,4,5,6},
xticklabels={log,sqrt,exp,cos,div,\(\displaystyle x^2\),mul},
xticklabels={5,3,5,10,5,2,2},
xtick style={color=black},
xticklabel style={color=color2},
y grid style={white!50.1960784313725!black},
ymin=-0.055, ymax=2,
ytick style={color=black}
]
\end{axis}

\end{tikzpicture}\label{fig:R2_p_freq}}
   \hfill
   \subfigure[S5 dataset]{\small% This file was created by tikzplotlib v0.9.1.
\begin{tikzpicture}

\definecolor{color0}{rgb}{0.12156862745098,0.466666666666667,0.705882352941177}
\definecolor{color1}{rgb}{1,0.498039215686275,0.0549019607843137}
\definecolor{color2}{rgb}{0.8, 0.39843137254902006, 0.04392156862745096}

\begin{axis}[
height=0.5\figheight,
legend cell align={left},
legend style={fill opacity=0.8, draw opacity=1, text opacity=1, at={(0.03,0.97)}, anchor=north west, draw=white!80!black},
tick pos=left,
width=0.5\figwidth,
x grid style={white!50.1960784313725!black},
xmin=-0.63, xmax=6.63,
xtick style={color=black},
xtick={0,1,2,3,4,5,6},
xticklabels={log,sqrt,exp,cos,div,\(\displaystyle x^2\),mul},
y grid style={white!50.1960784313725!black},
ylabel={[\%]},
ymin=0, ymax=42.4641840550506,
ytick style={color=black},
ytick={0,10,20,30,40,50},
yticklabels={\(\displaystyle 0\),\(\displaystyle 10\),\(\displaystyle 20\),\(\displaystyle 30\),\(\displaystyle 40\),\(\displaystyle 50\)},
]
\draw[draw=none,fill=color0,postaction={pattern=crosshatch dots}] (axis cs:-0.3,0) rectangle (axis cs:0,6.21120634558892);
\addlegendimage{ybar,ybar legend,draw=none,fill=color0,postaction={pattern=crosshatch dots}};
\addlegendentry{plain}

\draw[draw=none,fill=color0,postaction={pattern=crosshatch dots}] (axis cs:0.7,0) rectangle (axis cs:1,0.477758287599508);
\draw[draw=none,fill=color0,postaction={pattern=crosshatch dots}] (axis cs:1.7,0) rectangle (axis cs:2,8.55198063117332);
\draw[draw=none,fill=color0,postaction={pattern=crosshatch dots}] (axis cs:2.7,0) rectangle (axis cs:3,23.1675005822326);
\draw[draw=none,fill=color0,postaction={pattern=crosshatch dots}] (axis cs:3.7,0) rectangle (axis cs:4,16.632882136882);
\draw[draw=none,fill=color0,postaction={pattern=crosshatch dots}] (axis cs:4.7,0) rectangle (axis cs:5,34.2765508691984);
\draw[draw=none,fill=color0,postaction={pattern=crosshatch dots}] (axis cs:5.7,0) rectangle (axis cs:6,10.6821211473253);
\draw[draw=none,fill=color1,postaction={pattern=north east lines}] (axis cs:2.77555756156289e-17,0) rectangle (axis cs:0.3,5.84800122591077);
\addlegendimage{ybar,ybar legend,draw=none,fill=color1,postaction={pattern=north east lines}};
\addlegendentry{$c_{\cos}=5$}

\draw[draw=none,fill=color1,postaction={pattern=north east lines}] (axis cs:1,0) rectangle (axis cs:1.3,0.615662702477915);
\draw[draw=none,fill=color1,postaction={pattern=north east lines}] (axis cs:2,0) rectangle (axis cs:2.3,9.44898751474664);
\draw[draw=none,fill=color1,postaction={pattern=north east lines}] (axis cs:3,0) rectangle (axis cs:3.3,14.4216976214782);
\draw[draw=none,fill=color1,postaction={pattern=north east lines}] (axis cs:4,0) rectangle (axis cs:4.3,17.3381087865904);
\draw[draw=none,fill=color1,postaction={pattern=north east lines}] (axis cs:5,0) rectangle (axis cs:5.3,40.4420800524292);
\draw[draw=none,fill=color1,postaction={pattern=north east lines}] (axis cs:6,0) rectangle (axis cs:6.3,11.885462096367);
\end{axis}
\begin{axis}[
% axis x line=top,
axis y line=none,
height=0.5\figheight,
width=0.5\figwidth,
% log basis x={10},
tick pos=right,
x grid style={white!50.1960784313725!black},
% xmin=0.0616595001861482, xmax=16.2181009735893,
% xmode=log,
xlabel={$c_u$},
xmin=-0.63, xmax=6.63,
xtick style={color=black},
xtick={0,1,2,3,4,5,6},
xticklabels={log,sqrt,exp,cos,div,\(\displaystyle x^2\),mul},
xticklabels={1,1,1,5,1,1,1},
xtick style={color=black},
xticklabel style={color=color2},
y grid style={white!50.1960784313725!black},
ymin=-0.055, ymax=2,
ytick style={color=black}
]
\end{axis}

\end{tikzpicture}\label{fig:S5_p_freq}}
   % \hfill
   % \subfigure[A2 dataset]{\small\input{./fig/springer_2021-03-18/expr_analysis/Ndsr1_log_all_stats_pt}\label{fig:A2_p_freq}}
   \end{small}
    \caption{Relative frequency of atomic units with and without domain specific factors over all
    found equations
    (5 experiemnts, each with 78 different regularization strength, see appendix~\ref{app:ieql_training})
    of the combustion engine experiments in (a) and all found equations of the S5 dataset experiments in (b).
    Domain specific complexity factors $c_u$ are shown in the upper x-axis.  
    }
    \label{fig:R2_S5_p_freq}
\end{figure}

    %Georg: Why 390?
    \Figref{fig:R2_p_freq} shows the impact of motor specific complexity factors (shown in \tableref{ta:complexity_units})
    on the relative frequencies of atomic units evaluated on all found equations 
    (5 experiemnts, each with 78 different regularization strength, see appendix~\ref{app:ieql_training})
    on the real world dataset
    \textit{`torque of an internal combustion engine`}, which is introduced in \secref{sec:comb_engine}.
    The motor specific complexity factors clearly reduce the relative frequency of $\{\exp, \cos,/\}$
    atomic units and enhances polynomial structures with $\{x^2, *\}$ units. The $\sqrt{\phantom{x}}$ unit has been  barely chosen at all.
    This coincides with the expected behavior. % inferior??
    \Figref{fig:S5_p_freq} shows the impact of penalizing $\cos$ units with $(c_{\cos}=5)$
    on a more sophisticated synthetic dataset (S5), as expected the relative frequency of the
    cosine units is reduced.
    \subsection{Power Loss of an Electric Machine}\label{sec:power_loss}
%
    %%%%%%%%%%%%%%%%%%%%%%%%%%%%%%%%%%%%%%%%%%%%%%%%%%%%%%%%%%%%%%%%%%%%%%%%%%%%%%%%
    % data properties power loss
    %%%%%%%%%%%%%%%%%%%%%%%%%%%%%%%%%%%%%%%%%%%%%%%%%%%%%%%%%%%%%%%%%%%%%%%%%%%%%%%%

    %%%%%%%%%%%% reduced table version
    %%%%%%%% looks messy
    \begin{table}[t]
    \caption{
    Data properties for the power loss of an electric machine. The
    train domain covers just $80\%$ the range of operation. Except for rotor
    temperature $T_{\text{rot}}$ which has just three different operation points.
    }
        \vspace{-.5em}
    \label{ta:power_loss}
        % \scriptsize
        \small
        \begin{center}
        % \begin{tabular}{c@{\quad}c@{\quad\quad}c@{\quad}c@{\quad\quad}c@{\quad}c@{\quad}c@{\quad}c}
        \begin{tabular}{c@{\quad}c@{\quad}c@{\quad}c@{\quad}c}
        \toprule
                                  % & \multicolumn{2}{c}{Data Range}   & &\\
            Quantity             & Test Domain   & Train Domain                  &
    Description       & Type\\
            \midrule
    %        $I_{\text{D}}$       & A         & $-525$ & $-1$    & $-451$ & $-76$
            $I_{\text{D}}[\text{A}]$         & $[1,525]$    & $[76,451]$
    &
    direct current    & input  \\
            $I_{\text{Q}}[\text{ A}]$         & $[1,525]$   & $[76,451]$
    &
    quadratic current & input  \\
            $T_{\text{rot}}[^\circ\text{C}]$ & $[-20,150]$   & $[-20,150]$
    &
    rotor temperature & input  \\
            $n_{\text{mtr}}[\text{rpm}]$       & $[1,16000]$ & $[2001,14001]$
    &
    motor speed       & input  \\
            \midrule

            $P_{\text{mod}}[\text{W}]$         & $[0,4558]$  & $[95,3364]$
    &
    power loss        & output \\
            \bottomrule
        \end{tabular}
        \end{center}
    \end{table}
    In this section, we address with the power loss of an electric machine a
    relevant problem in the industrial domain~\citep{Buchneretal2020}.
    Increasing requirements on control of electric machines in automotive
    power trains lead to the necessity of more and more sophisticated and complex
    models to describe the system.
    Those models have to be able to capture non-linear effects
    at different components in the system, as in the dataset for the power loss of an electric machine.
    Further, those models have to be minimal in computational effort and
    memory demand due to embedded hardware and latency constraints in the range of micro-seconds.

    In this industrial dataset, the power loss of a winding head is measured
    depending on the
     direct current      $I_{\text{D}}$,
     quadratic current   $I_{\text{Q}}$,
     rotor temperature   $T_{\text{rot}}$,
     and
     motor speed         $n_{\text{mtr}}$.
    The data was measured at stationary operation points.
    It contains $24684$ data points measured at equidistant variations of the
    quantities listed in \tableref{ta:power_loss} within their
    range of operation.
    The test dataset contains unseen data from the entire domain and training dataset
    contains data only from $80\%$ of its range of operation.
    Further details on data preparation are
    given in appendix~\ref{app:power_loss_data}.

    We use the model selection criteria \ValISparse on the found equations of \bEQL, \dEQL and \pysr.
    \begin{table}[t]
        \caption{Results on real-world datasets.
          Reported are median, minimum and maximum (in sub and superscript)
          of root mean squared error (RMSE) on the real world test datasets.
          Domain knowledge is used for \bEQLmotor as given in \tableref{ta:complexity_units}.
          We use the model selection criteria \ValISparse and state the number of active parameters.
          For the combustion engine dataset we also present the best validation models selected with \ValI.
        }
        % \vspace{-.5em}
        \label{ta:etas}
        % \scriptsize
        \centering
        \begin{adjustbox}{max width=\linewidth}
        \begin{tabular}{@{\ }l@{}r@{}lr@{}l@{}|r@{}lr@{}l|r@{}lr@{}l}%cc@{\ }}
\toprule
    {} &\multicolumn{4}{c|}{electric machine [W]}& \multicolumn{6}{c}{combustion engine [Nm]} \\
         {} &  \multicolumn{2}{c}{ \ValISparse}  &\multicolumn{2}{c|}{\#param.} &   \multicolumn{2}{c}{ \ValISparse}  &\multicolumn{2}{c}{\#param.}  &        \multicolumn{2}{c}{ \ValI}  & \multicolumn{2}{c}{\#param.}\\
\midrule                                                                                        
%    MP         &                        1042.70&&--&&               60.17&              &                  --            & &-- &&--         \\
%    \pysr      &      188.40 &\mm{155.84}{347.93} & 9  &\mm{3}{9} &  16.79&\mm{13.35}{22.37}  &7& \mm{5}{11}            & 5.91& \mm{5.00}{7.66}    &  33 &\mm{23}{40}  \\
%%    GP all train         &                           9.69&&& &                1.17&              &                  --            &      \\
%    GP         &                           9.69&&--& &                2.14&           &                  --            &&--   &&--   \\
%% MLP        &  \mmm{97.98}{107.98}{130.68}& &  \mmm{1.50}{1.79}{1.9           } &  \mmm{1.50}{1.79}{1.900}
%    \dEQL      &       0.03  &\mm{0.03}{0.04}     & 10 &\mm{7}{14} &  1.75&\mm{1.55}{1.96}&     95&\mm{70}{129}         & 1.46& \mm{1.35}{1.61}    &  382& \mm{225}{427} \\
%    \bEQL      &       0.03  &\mm{0.02}{0.03}     & 6  &\mm{6}{6} &   2.48&\mm{2.18}{3.04}&     65&\mm{42}{79}          & 1.44& \mm{1.40}{1.71}    &  448& \mm{395}{470}  \\
%    \bEQLmotor &       0.02  &\mm{0.02}{0.02}     & 6  &\mm{6}{6} &   3.17&\mm{2.90}{6.16}&     32&\mm{15}{48}          & 1.60& \mm{1.39}{1.79}    &  339& \mm{170}{386}  \\
%    \bottomrule
%\midrule
    MP         & 1042.70&                      &  &                 & 60.17&                            &  &                 & 60.17  &                   &  &                   \\
    GP         & 0.92&                         &  &                & 1.79   &                          &    &               & 1.79      &               &     &                \\
% GP all train     & 9.69&                         &  &                & 1.17   &                          &    &               & --                     &                     \\
% MLP      & \mmm{97.98}{107.98}{130.68}  &                 & \mmm{1.50}{1.79}{1.9           } & \mmm{1.50}{1.79}{1.900}
\pysrA     & \mmmt{155.84}{188.40}{347.93} & \mmmt{3}{9}{9}   & \mmmt{13.35}{16.79}{22.37}        & \mmmt{5}{7}{11}    & \mmmt{5.00}{5.91}{7.66} & \mmmt{23}{33}{40}    \\
\pysrB     & \mmmt{230.23}{230.23}{230.23} & \mmmt{8}{8}{8}    &\mmmt{16.79}{22.37}{22.37}        &\mmmt{5}{5}{7}      &\mmmt{4.86}{5.33}{6.29} &\mmmt{34}{35}{39} \\
\dEQL      & \mmmt{0.03}{0.03}{0.04}       & \mmmt{7}{10}{14} & \mmmt{1.55}{1.75}{1.96}           & \mmmt{70}{95}{129} & \mmmt{1.35}{1.46}{1.61} & \mmmt{225}{382}{427} \\
\bEQL      & \mmmt{0.02}{0.03}{0.03}       & \mmmt{6}{6}{6}   & \mmmt{2.18}{2.48}{3.04}           & \mmmt{42}{65}{79}  & \mmmt{1.40}{1.44}{1.71} & \mmmt{395}{448}{470} \\
\bEQLmotor & \mmmt{0.02}{0.02}{0.02}       & \mmmt{6}{6}{6}   & \mmmt{2.90}{3.17}{6.16}           & \mmmt{15}{32}{48}  & \mmmt{1.39}{1.60}{1.79} & \mmmt{170}{339}{386} \\
\bottomrule
\end{tabular}
% pysrA PE pysr criteria:  \mmm{0.02}{0.47}{159.11} complexity: \mmm{9}{19}{23} optimal results with complexit 15-19 
% pysrB PE VInt: \mmm{0.03}{0.03}{0.03}, complexity: \mmm{37}{39}{40}
% pysrB PE pysr criteria:  \mmm{0.02}{0.02}{0.16} complexity: \mmm{15}{17}{19} optimal results with complexit 15-19 
% pysr R2 pysr criteria \mmm{16.79}{16.79}{22.37}, complexity: \mmm{5}{7}{7}:

        \end{adjustbox}
    \end{table}
    %%%%%%%%%%%%%%%%%%%%%%%%%%%%%%%%%%%%%%%%%%%%%%%%%%%%%%%%%%%%%%%%%%%%%%%%%%%%%%%%
Results are shown in \tableref{ta:etas}.
\bEQL and \dEQL outperform the genetic algorithm (GA) and the Gaussian Process (GP) even on the training data and provide very
accurate predictions on the test set.
The GP does not capture the underlying relationship as can be seen at the
large test RMSE.
A closer look at the results of the genetic algorithm revealed
that the internal selection criteria of \pysr led to competitive results:
\pysrA with \mmm{0.02}{0.47}{159.11}W and \mmm{9}{19}{23}parameters 
and \pysrB with \mmm{0.02}{0.02}{0.16}W and \mmm{15}{17}{19}parameters.
\bEQL needs even fewer parameters than \dEQL.
This is due to the use of copy units, which allow to reusing features from previous layers directly.
We emphasize that, out of the large set of possible functions, the \bEQL reliably extracted
a simple quadratic equation that suitably describes the dataset.
To highlight the simplicity of the \bEQL's result, we state in the following
the structure of the selected equation, which is the same for all 5 experiments
\begin{align}\label{eq:PE_expr}
    y=& w_1 I_Q + w_2 I_D +w_5 (w_3 I_D+b_1)^{2}  +w_6 (w_4 I_Q+ b_2)^{2} + b_3\,.
\end{align}
Weights are indicated with $w$ and bias with $b$. 
All selected equations can be simplified to the same equation
$y =  0.4 I_D^2 - 1.12 I_D + 0.41 I_Q^2 + 1.13 I_Q - 0.31$.
Numbers were rounded to three figures and input and
output dimensions were anonymized.
In addition, we emphasize that the algorithm learned that only two of the input
variables, namely $[I_D,\, I_Q]$, are relevant to the output value, while the others
have negligible influence.

\subsection{Torque Model of an Internal Combustion Engine}\label{sec:comb_engine}
% table data properties torque model
%%%%%%%%%%%%%%%%%%%%%%%%%%%%%%%%%%%%%%%%%%%%%%%%%%%%%%%%%%%%%%%%%%%%%%%%%%%%%%%%
\begin{table}[tb]
\caption{%
Data properties for the torque model of an internal combustion engine.%Georg: why internal? --> technical name for Verbrennungsmotor, it's also correct without 'internal'
}\label{ta:comb_engine}
% \vspace*{-.9em}
\centering
\begin{adjustbox}{max width=\linewidth}
    \begin{tabular}{@{}c@{\ \ }c@{\quad}r@{\quad}r@{\quad}l@{\quad}c@{}}
        \toprule
        Quantity             & Unit      & Min.   & Max.     & Description

& Type\\
        \midrule
        $\phi_{\text{ex}}$   & $^\circ$Crank & $-20$     & $20$    & exhaust
camshaft & input  \\
        $\phi_{\text{in}}$   & $^\circ$Crank & $-4$     & $36$    & intake
camshaft  & input  \\
        $r_{\text{l}}$       & $\%$          & $13$     & $86$    & relative
load    & input  \\
        $\phi_{\text{ign}}$  & $^\circ$Crank & $-27$    & $61$    & ignition
angle   & input  \\
        $n_{\text{eng}}$     & rpm           & $597$    & $6000$  & engine
speed
    & input  \\
        \midrule
        $M_{\text{eng}}$     & Nm            & $-38$    & $261$   & engine
torque    & output \\
        \bottomrule
    \end{tabular}
    \end{adjustbox}
\end{table}
%%%%%%%%%%%%%%%%%%%%%%%%%%%%%%%%%%%%%%%%%%%%%%%%%%%%%%%%%%%%%%%%%%%%%%%%%%%%%%%%
A second task from the industrial domain is the modeling of the torque of a
combustion engine in dependence of the control parameters.
Ongoing improvements in power train technologies
require more and more precise models for control strategies.
This is commonly correlated with a complex structure of the models as well as
the large number of calibration parameters.
For utilization of those models in embedded systems it is essential
to have fast development cycles, \ie an automated adaptation to a new system,
 and low production costs.
The unit costs grow with the computational needs on an embedded controller, thus
computationally efficient at evaluation time is important.
In practice, this leads often to a trade-off between accuracy of the model and its complexity.

In this dataset the engine torque is measured depending on
 exhaust camshaft $\phi_{\text{ex}}$,
intake camshaft   $\phi_{\text{in}}$,
 relative load   $r_{\text{l}}$,
 ignition angle  $\phi_{\text{ign}}$
and
 engine speed    $n_{\text{eng}}$.
The data was measured at stationary operation points. It contains $1775$ data points measured at
 variations of the quantities listed in \tableref{ta:comb_engine} within their
range of operation.
The dataset is split into $80\%$ training and $20\%$ testing data and $10\%$ of the train data set is used for validation.
\begin{figure}[t]
    \begin{small}
   \centering
   \subfigure[pareto plot]{\small% This file was created by tikzplotlib v0.9.1.
\begin{tikzpicture}

\definecolor{color0}{rgb}{0.12156862745098,0.466666666666667,0.705882352941177}
\definecolor{color1}{rgb}{0.580392156862745,0.403921568627451,0.741176470588235}

\begin{axis}[
height=0.5\figheight,
legend cell align={left},
legend style={kurze Legende},
legend style={font=\scriptsize,fill opacity=0.5, draw opacity=1, text opacity=1, draw=white!80!black},
log basis x={10},
log basis y={10},
tick pos=left,
width=0.5\figwidth,
x grid style={white!50.1960784313725!black},
xlabel={complexity},
xmin=3.8602947784138, xmax=1143.6950423289,
xmode=log,
xtick style={color=black},
xtick={0.1,1,10,100,1000,10000,100000},
xticklabels={\(\displaystyle 10^{-1}\),\(\displaystyle 10^{0}\),\(\displaystyle 10^{1}\),\(\displaystyle 10^{2}\),\(\displaystyle 10^{3}\),\(\displaystyle 10^{4}\),\(\displaystyle 10^{5}\)},
y grid style={white!50.1960784313725!black},
ylabel={RMSE [Nm]},
ymin=1.44082702149181, ymax=14.0295163030217,
ymode=log,
ytick style={color=black},
ytick={2, 4, 10},
yticklabels={\(\displaystyle 2\),\(\displaystyle 4\),\(\displaystyle 10\)}%,\(\displaystyle 10^{3}\)}
]
\addplot [semithick, color0, mark=*, mark size=2, mark options={solid}, only marks]
table {%
5 7.0753657382539
5 12.6506844440411
5 7.046988129897
6 5.29608745321251
6 5.3068195666099
7 7.60944736508441
8 5.52193930337631
9 5.28208328769929
9 5.48239943353095
10 5.37446158269612
12 5.49530031481263
15 5.65056093877338
16 5.32790922995368
18 7.03569471566374
18 5.04226082556938
19 5.31181363302154
20 5.45701900297791
22 5.3447567115352
22 4.64798757070753
28 4.10883720337547
36 3.74207605520185
37 3.39962249070339
37 3.97803924491523
40 3.42098825152036
47 3.8607236043276
48 3.71990045947844
52 2.89582713553928
52 3.74139971560375
55 2.87751744537703
63 2.99370149786672
64 3.08699629618144
70 3.84823299481381
70 3.11295869985414
72 3.51038629464526
75 3.39551795300332
83 2.76415323233946
97 3.1282886557638
109 2.57780588678149
122 2.72752353843523
125 2.86423530996526
131 2.83553500764174
132 2.3818476103362
141 2.29973575390919
154 2.37395546191235
170 2.34472684133176
183 1.93150298708562
186 2.28053860657136
207 2.5821460254189
218 2.346851116574
221 2.54033486180173
235 1.87506898542667
256 2.08749315262442
256 2.2657867311934
272 2.08709290198762
323 2.13254917450361
351 2.35164851136173
362 2.14227448309994
387 1.59786660376113
438 2.00584544569563
464 2.4185935834009
487 1.61954839334255
497 2.03052635650355
498 2.31684509203192
522 1.97130565518717
528 1.87098114550589
541 2.1241950817924
546 1.78342702672419
605 3.28545515914101
630 2.51910233490204
654 1.93427555997043
751 1.69526145886796
756 3.76056545448667
804 2.37044367307784
838 1.70986648864382
845 2.17833940746895
859 2.15054027236039
871 1.81532505198822
883 1.75496152237604
};
\addlegendentry{validation}
\addplot [ultra thick, color1, dotted]%, forget plot]
table {%
55 1.44082702149181
55 14.0295163030217
};
\addlegendentry{s.eq}
\addplot [ultra thick, white!49.8039215686275!black, dash pattern=on 1pt off 3pt on 3pt off 3pt]%, forget plot]
table {%
387 1.44082702149181
387 14.0295163030217
};
\addlegendentry{eq-1}
\end{axis}

\end{tikzpicture}\label{fig:R2_pareto}}
   \hfill
   \subfigure[histogram of residuals]{\small% This file was created by tikzplotlib v0.9.1.
\begin{tikzpicture}

\definecolor{color0}{rgb}{0.580392156862745,0.403921568627451,0.741176470588235}

\begin{axis}[
height=0.5\figheight,
legend cell align={left},
legend style={fill opacity=0.5, draw opacity=1, text opacity=1, draw=white!80!black},
tick pos=left,
width=0.5\figwidth,
x grid style={white!50.1960784313725!black},
xlabel={residuals [Nm]},
xmin=-11, xmax=11,
xtick style={color=black},
xtick={-15,-10,-5,0,5,10,15},
xticklabels={\(\displaystyle -15\),\(\displaystyle -10\),\(\displaystyle -5\),\(\displaystyle 0\),\(\displaystyle 5\),\(\displaystyle 10\),\(\displaystyle 15\)},
ymajorticks=false, ymin=0, ymax=0.290677966101695
]
\draw[draw=none,fill=color0,fill opacity=0.6,postaction={pattern=crosshatch dots}] (axis cs:-10,0) rectangle (axis cs:-9,0.00568181818181818);
\addlegendimage{ybar,ybar legend,draw=none,fill=color0,fill opacity=0.6,postaction={pattern=crosshatch dots}};
\addlegendentry{s.eq}

\draw[draw=none,fill=color0,fill opacity=0.6,postaction={pattern=crosshatch dots}] (axis cs:-9,0) rectangle (axis cs:-8,0.00568181818181818);
\draw[draw=none,fill=color0,fill opacity=0.6,postaction={pattern=crosshatch dots}] (axis cs:-8,0) rectangle (axis cs:-7,0.00284090909090909);
\draw[draw=none,fill=color0,fill opacity=0.6,postaction={pattern=crosshatch dots}] (axis cs:-7,0) rectangle (axis cs:-6,0.0142045454545455);
\draw[draw=none,fill=color0,fill opacity=0.6,postaction={pattern=crosshatch dots}] (axis cs:-6,0) rectangle (axis cs:-5,0.0170454545454545);
\draw[draw=none,fill=color0,fill opacity=0.6,postaction={pattern=crosshatch dots}] (axis cs:-5,0) rectangle (axis cs:-4,0.0426136363636364);
\draw[draw=none,fill=color0,fill opacity=0.6,postaction={pattern=crosshatch dots}] (axis cs:-4,0) rectangle (axis cs:-3,0.0511363636363636);
\draw[draw=none,fill=color0,fill opacity=0.6,postaction={pattern=crosshatch dots}] (axis cs:-3,0) rectangle (axis cs:-2,0.0767045454545455);
\draw[draw=none,fill=color0,fill opacity=0.6,postaction={pattern=crosshatch dots}] (axis cs:-2,0) rectangle (axis cs:-1,0.136363636363636);
\draw[draw=none,fill=color0,fill opacity=0.6,postaction={pattern=crosshatch dots}] (axis cs:-1,0) rectangle (axis cs:0,0.15625);
\draw[draw=none,fill=color0,fill opacity=0.6,postaction={pattern=crosshatch dots}] (axis cs:0,0) rectangle (axis cs:1,0.110795454545455);
\draw[draw=none,fill=color0,fill opacity=0.6,postaction={pattern=crosshatch dots}] (axis cs:1,0) rectangle (axis cs:2,0.127840909090909);
\draw[draw=none,fill=color0,fill opacity=0.6,postaction={pattern=crosshatch dots}] (axis cs:2,0) rectangle (axis cs:3,0.107954545454545);
\draw[draw=none,fill=color0,fill opacity=0.6,postaction={pattern=crosshatch dots}] (axis cs:3,0) rectangle (axis cs:4,0.0738636363636364);
\draw[draw=none,fill=color0,fill opacity=0.6,postaction={pattern=crosshatch dots}] (axis cs:4,0) rectangle (axis cs:5,0.0482954545454545);
\draw[draw=none,fill=color0,fill opacity=0.6,postaction={pattern=crosshatch dots}] (axis cs:5,0) rectangle (axis cs:6,0.0113636363636364);
\draw[draw=none,fill=color0,fill opacity=0.6,postaction={pattern=crosshatch dots}] (axis cs:6,0) rectangle (axis cs:7,0.00568181818181818);
\draw[draw=none,fill=color0,fill opacity=0.6,postaction={pattern=crosshatch dots}] (axis cs:7,0) rectangle (axis cs:8,0.00568181818181818);
\draw[draw=none,fill=color0,fill opacity=0.6,postaction={pattern=crosshatch dots}] (axis cs:8,0) rectangle (axis cs:9,0);
\draw[draw=none,fill=color0,fill opacity=0.6,postaction={pattern=crosshatch dots}] (axis cs:9,0) rectangle (axis cs:10,0);
\draw[draw=none,fill=white!49.8039215686275!black,fill opacity=0.6,postaction={pattern=north east lines}] (axis cs:-10,0) rectangle (axis cs:-9,0);
\addlegendimage{ybar,ybar legend,draw=none,fill=white!49.8039215686275!black,fill opacity=0.6,postaction={pattern=north east lines}};
\addlegendentry{eq-1}

\draw[draw=none,fill=white!49.8039215686275!black,fill opacity=0.6,postaction={pattern=north east lines}] (axis cs:-9,0) rectangle (axis cs:-8,0.00282485875706215);
\draw[draw=none,fill=white!49.8039215686275!black,fill opacity=0.6,postaction={pattern=north east lines}] (axis cs:-8,0) rectangle (axis cs:-7,0);
\draw[draw=none,fill=white!49.8039215686275!black,fill opacity=0.6,postaction={pattern=north east lines}] (axis cs:-7,0) rectangle (axis cs:-6,0);
\draw[draw=none,fill=white!49.8039215686275!black,fill opacity=0.6,postaction={pattern=north east lines}] (axis cs:-6,0) rectangle (axis cs:-5,0.00282485875706215);
\draw[draw=none,fill=white!49.8039215686275!black,fill opacity=0.6,postaction={pattern=north east lines}] (axis cs:-5,0) rectangle (axis cs:-4,0.0169491525423729);
\draw[draw=none,fill=white!49.8039215686275!black,fill opacity=0.6,postaction={pattern=north east lines}] (axis cs:-4,0) rectangle (axis cs:-3,0.0169491525423729);
\draw[draw=none,fill=white!49.8039215686275!black,fill opacity=0.6,postaction={pattern=north east lines}] (axis cs:-3,0) rectangle (axis cs:-2,0.0423728813559322);
\draw[draw=none,fill=white!49.8039215686275!black,fill opacity=0.6,postaction={pattern=north east lines}] (axis cs:-2,0) rectangle (axis cs:-1,0.169491525423729);
\draw[draw=none,fill=white!49.8039215686275!black,fill opacity=0.6,postaction={pattern=north east lines}] (axis cs:-1,0) rectangle (axis cs:0,0.242937853107345);
\draw[draw=none,fill=white!49.8039215686275!black,fill opacity=0.6,postaction={pattern=north east lines}] (axis cs:0,0) rectangle (axis cs:1,0.27683615819209);
\draw[draw=none,fill=white!49.8039215686275!black,fill opacity=0.6,postaction={pattern=north east lines}] (axis cs:1,0) rectangle (axis cs:2,0.138418079096045);
\draw[draw=none,fill=white!49.8039215686275!black,fill opacity=0.6,postaction={pattern=north east lines}] (axis cs:2,0) rectangle (axis cs:3,0.0593220338983051);
\draw[draw=none,fill=white!49.8039215686275!black,fill opacity=0.6,postaction={pattern=north east lines}] (axis cs:3,0) rectangle (axis cs:4,0.0112994350282486);
\draw[draw=none,fill=white!49.8039215686275!black,fill opacity=0.6,postaction={pattern=north east lines}] (axis cs:4,0) rectangle (axis cs:5,0.0112994350282486);
\draw[draw=none,fill=white!49.8039215686275!black,fill opacity=0.6,postaction={pattern=north east lines}] (axis cs:5,0) rectangle (axis cs:6,0.00282485875706215);
\draw[draw=none,fill=white!49.8039215686275!black,fill opacity=0.6,postaction={pattern=north east lines}] (axis cs:6,0) rectangle (axis cs:7,0.00564971751412429);
\draw[draw=none,fill=white!49.8039215686275!black,fill opacity=0.6,postaction={pattern=north east lines}] (axis cs:7,0) rectangle (axis cs:8,0);
\draw[draw=none,fill=white!49.8039215686275!black,fill opacity=0.6,postaction={pattern=north east lines}] (axis cs:8,0) rectangle (axis cs:9,0);
\draw[draw=none,fill=white!49.8039215686275!black,fill opacity=0.6,postaction={pattern=north east lines}] (axis cs:9,0) rectangle (axis cs:10,0);
\end{axis}

\end{tikzpicture}\label{fig:R2_hist}}
%    \vspace*{-.8em}
   \end{small}
    \caption{%
        Equation learning with \bEQLmotor on a real world dataset for torque of a combustion engine. Here, we show the
        run with median performance with the \ValISparse criteria.
        Panel (a) shows the pareto plot of the \bEQL
        and (b) shows the histogram of residuals of the equation (s.eq) selected with the \ValISparse criteria
        and a more complex equation (eq-1).
    }
    \label{fig:R2_plots}
\end{figure}
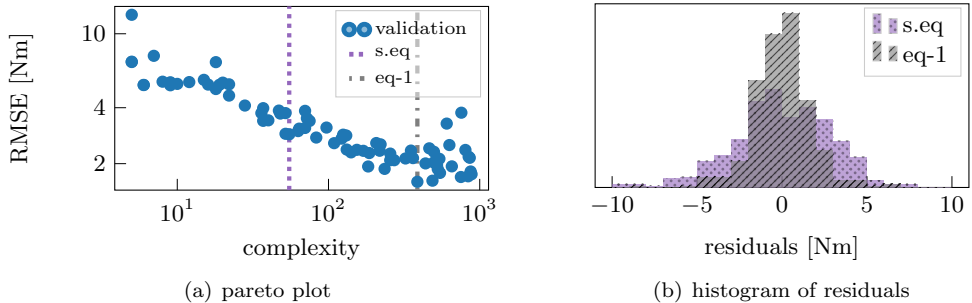

Instance selection can be done with the \ValISparse criteria or solely based on validation performance as long
as the equation is computationally efficient at evaluation time.
We present both solutions in \tableref{ta:etas} alongside with the number of active parameters.
The GP is supposed to perform best on this dataset, but the EQL-networks, selected with \ValI,
perform similar or even better. 
With about 170-470 parameters they are fast to compute and small enough to fit on embedded systems.
The \dEQL finds the best test RMSE results.
This might indicate that
for this dataset it is sufficient to deal with $\{\sin, \cos, *, \text{identity}\}$ units.
Despite its larger expressive power, 
\bEQL can compete with the \dEQL architecture.
\pysr is not able to learn equations that compete on the performance level, but it can identify
simple equations with about 33 parameters. Yet, \bEQLmotor also finds equations of similar complexity,
but with a twice as good median test RMSE.
\Tableref{ta:etas} shows a noticeable relation between the performance of the models and their complexity for \bEQL, \bEQLmotor and \dEQL.
The more complex the model the betters its test RMSE.
This effect occurs also in \figref{fig:R2_pareto}, which shows the pareto plot for the \bEQLmotor experiment (median
performance and \ValISparse criteria). The histograms of residuals of the selected equation
alongside with the best performing equation show no significant distribution shifts, see \figref{fig:R2_hist}.

We analyze the equations of
median performance and the equations with the smallest number of parameters
selected with the \ValISparse criteria for \bEQL and \bEQLmotor in appendix~\ref{app:R2_eq}.
Their test RMSE is not as good as a GP, but their structure is interpretable.
The motor specific complexity factors lead to equations of
mainly polynomial structure with higher degree than without motor specific complexity factors.
The latter combine polynomials with nonlinear units like $\cos,\exp, \log$ or division units.%Georg: why simple// %matthias:I removed it
Using more nonlinear units reduces the degree of the used polynomials.
This is in agreement with the analysis of relative frequencies of atomic units shown in \figref{fig:R2_p_freq}, which clearly shows that the use of motor specific complexity factors leads to a
reduction of the use of $\{\log,\cos,\exp\}$ and division units and an increase of $x^2$ and multiplication units.

\section{Conclusion}
\label{sec:conclusion}
We introduced the informed equation learner \bEQL. It aims to learn compact and interpretable models in form of concise
mathematical equations from data.
Our method differs from previous work in two key aspects: First, it is able to handle functional units with singularities
in all hidden layers.
Secondly, it can incorporate expert knowledge about the underlying system.
To do so, prior knowledge is encoded in a complexity cost for individual atomic units.
We evaluated our method by finding compact functional expressions for four different use cases:
The \bEQL performs well on established reference datasets. %~\citep{Martius2017}.
% A second use case demonstrated the use of customized priors to prefer certain functional terms in an
Customized priors can be applied to prefer certain functional terms in an
ambiguous dataset supporting several possible explanations.
Finally, we evaluated the algorithm on two real industrial problems.
In both settings, the underlying relations of the system were unknown a priori,
and the \bEQL was able to extract simple, explainable models describing the respective output variable.
Depending on the needs it also provides a high-performance solution with low computational cost.

\newpage
\begin{description}
    \item[Funding:]
    Philipp Hennig and Georg Martius are members of the Machine Learning
Cluster of Excellence, funded by the Deutsche Forschungsgemeinschaft
(DFG, German Research Foundation) under Germany’s Excellence Strategy
– EXC number 2064/1 – Project number 390727645.
We acknowledge the support from the German Federal Ministry of
Education and Research (BMBF) through the Tübingen AI Center (FKZ:
01IS18039B).
\item[Authors' contributions:]
All authors contributed to the conception and design of the study.
Code preparation, data collection and analysis were performed by MW
and supervised by GM and PH. AJ has supported the work as domain expert.
The manuscript was drafted by MW and edited and reviewed by all authors.
\end{description}

\bibliographystyle{abbrvnat}      % basic style, author-year citations
\bibliography{paper_arxiv}
%%%%%%%%%%%%%%%%%%%%%%%%%%%%%%%%%%%%%%%%%%%%%%%%%%%%%%%%%%%%%%%%%%%%%%%%%%%%%%
\newpage
\appendix
\section{Appendix}
\begin{figure}
    \subfigure[S0]  {\includegraphics[width=0.32\textwidth]{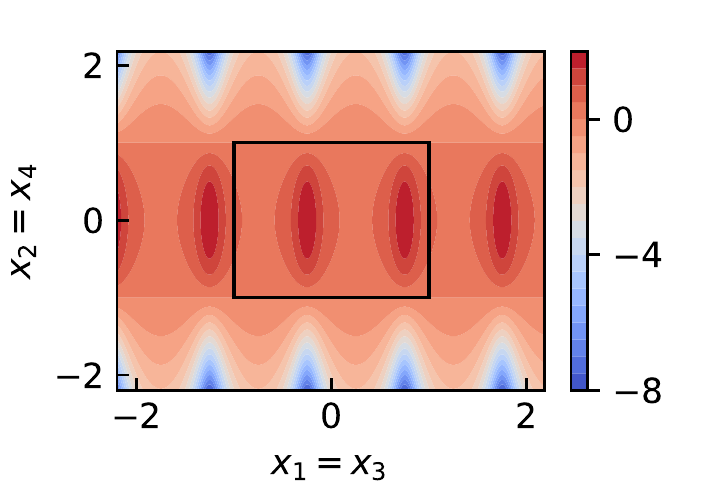}}
    \subfigure[S1]  {\includegraphics[width=0.32\textwidth]{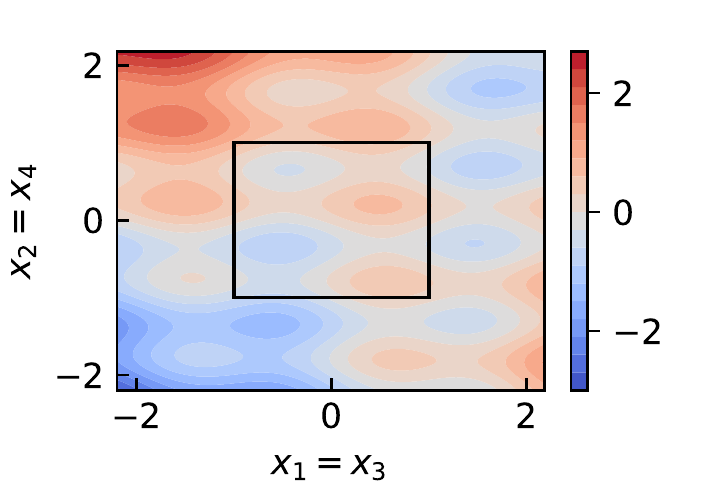}}
    \subfigure[S2]  {\includegraphics[width=0.32\textwidth]{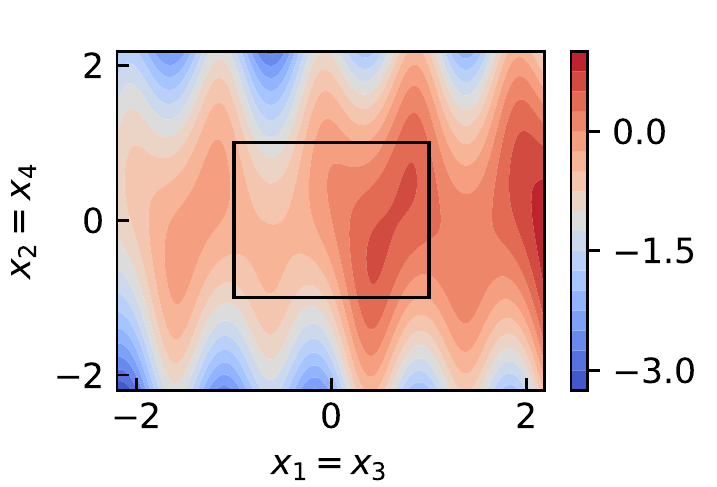}}
    \subfigure[S3]  {\includegraphics[width=0.32\textwidth]{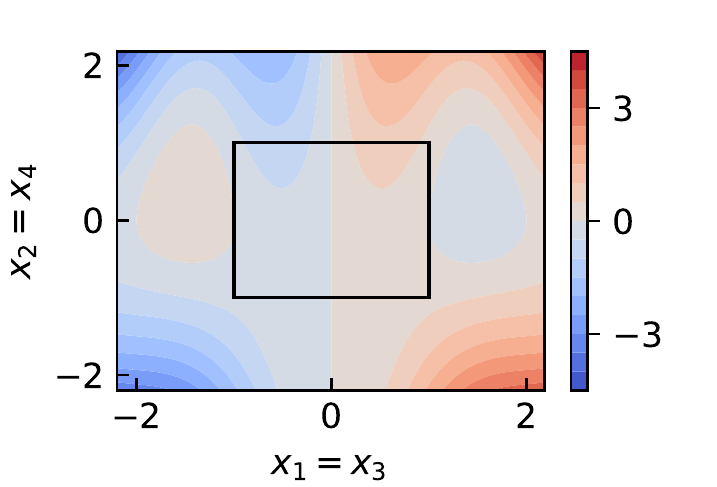}}
    \subfigure[S4]  {\includegraphics[width=0.32\textwidth]{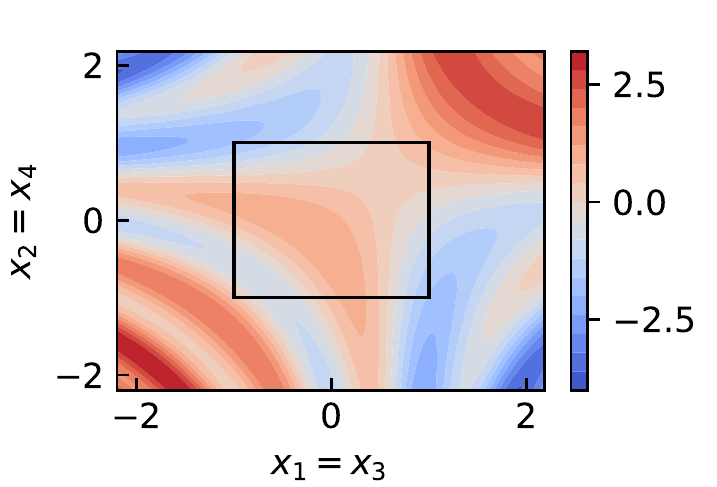}}
    \subfigure[S5]  {\includegraphics[width=0.32\textwidth]{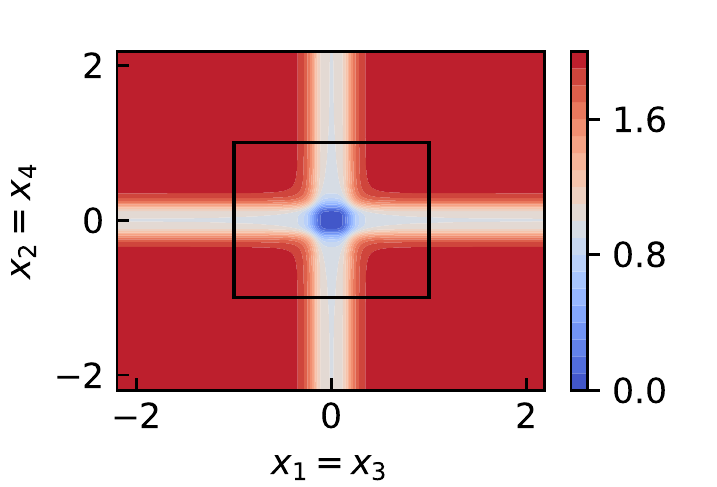}}
    \subfigure[S6]  {\includegraphics[width=0.32\textwidth]{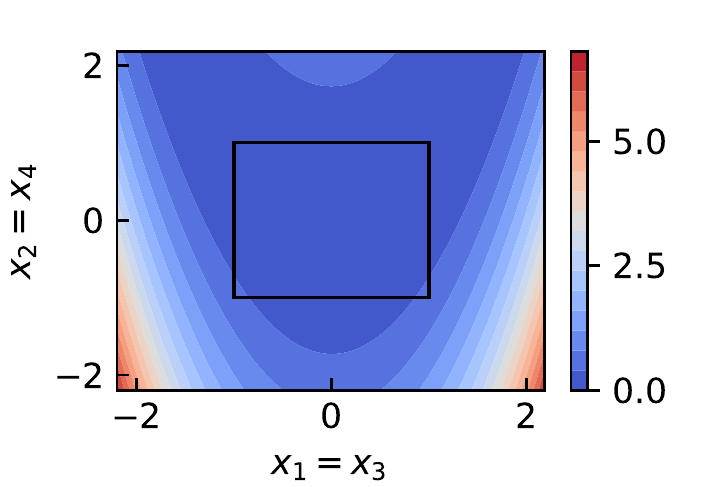}}
    \subfigure[A1]  {\includegraphics[width=0.32\textwidth]{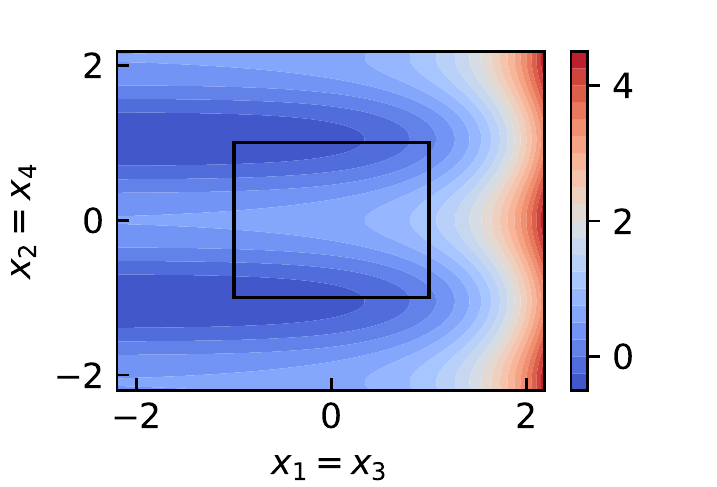}}
    \subfigure[A2]  {\includegraphics[width=0.32\textwidth]{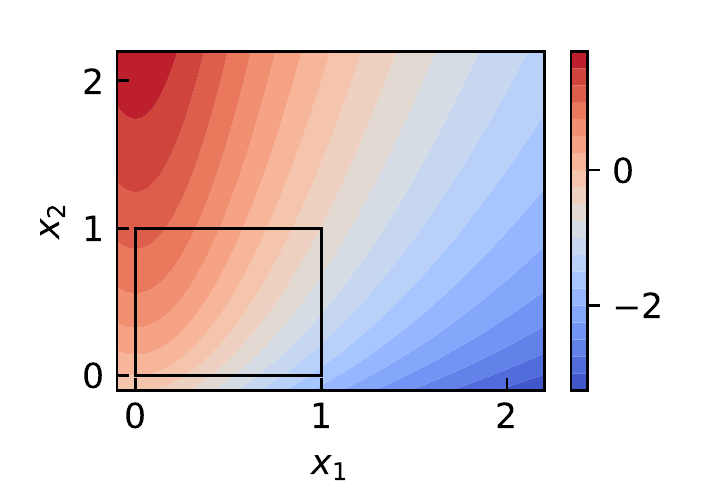}}
    \subfigure[A3]  {\includegraphics[width=0.32\textwidth]{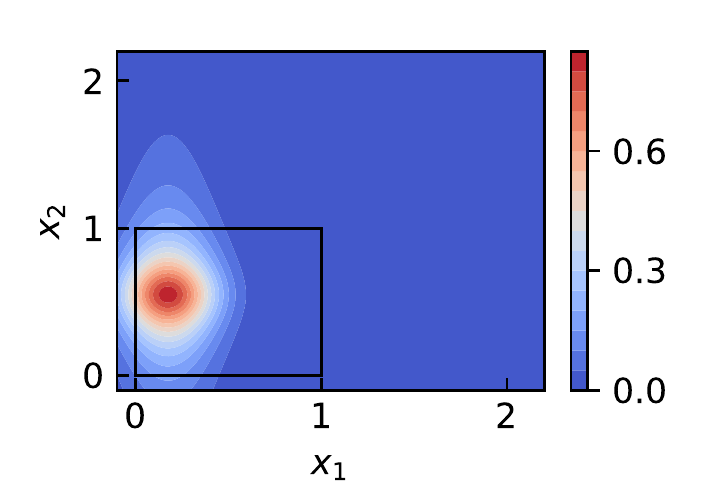}}
    \subfigure[A4]  {\includegraphics[width=0.32\textwidth]{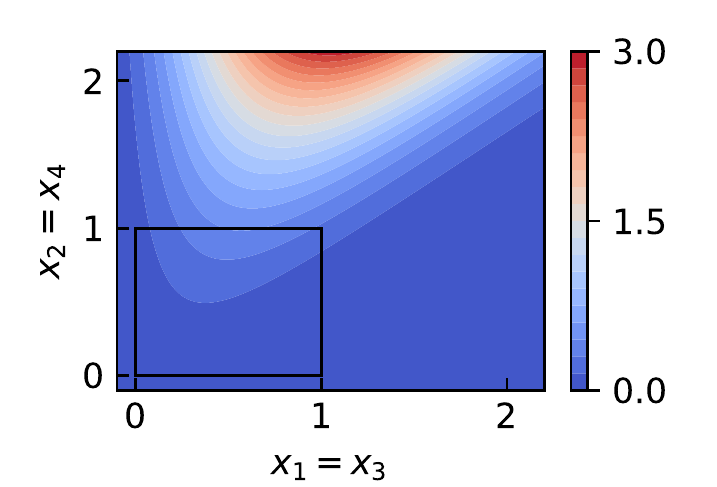}}
    \caption{Two dimensional slices through the input space of simulated equations S0--S6 and A1--A4 from \tableref{ta:datasets}.
    The black rectangle indicates the train domain.% $[-1,1]^4$.
    }
    \label{fig:dataset_contour}
\end{figure}
% \matthias{shall I move the Training \secref{sec:ieql_training} at this position?}
\subsection{\bEQL Training}\label{app:ieql_training}
All experiments are performed
using an \bEQL with four hidden layers.
Each hidden layer has
\(\{\cos,\,\exp{\;},\,\log,\, \sqrt{\phantom{x}},\,x^2,\, *,\,/\}\)
as atomic units, and each atomic unit is applied four times in each layer.
We prohibit the following combinations
$\cos(\cos)$, $\cos(\exp)$, $\exp(\exp)$,
    $\log(\log)$.
The \bEQL has thus $6405$ learnable weights.
% by the masking process described above
Training is executed in two phases.
In phase 1 we train for $T_1=2000$ epochs without regularization.
This phase overcomes bad initialization
and assures that the model is close to a minimum when pruning starts.
In phase 2 we train for $T_2=10000$ epochs with regularization strength $\lambda$.
Just for the combustion engine dataset, which has about 5 to 6 times fewer data points,
we increased the number of epochs by a factor of 8.
We skipped train phase 1 for the ambiguous dataset since the \bEQL
is already sufficiently close to a minimum.
After each epoch, an intrinsic penalty epoch is calculated
with $100$ randomly sampled data points from the test domain (without labels).
The maximum desired output value is set to $B=10$ with $\eta=1$.
    We use the \textsc{Adam} optimizer \citep{Kingma2015} with learning rate $0.001$, moving average $\beta_1=0.4$ and $\eps=10^{-8}$ for numerical
    stability.
The initial dropout rate of the Bernoulli gates is set to $0.5$
and the domain/bound penalty strength is set to $\delta=1$.
Further information on dropout rate is given in appendix~\ref{app:l0_method}.
Since the optimal regularization scale parameter $\lambda$ is not known in
advance, we train several models with different regularization strengths
$\lambda=10^k$ where $k$ is in the range from $-5.0$ to $0.0$ with 78 equally spaced steps.
This results in 78 equations with different
complexities and root mean squared errors (RMSE).

\subsubsection*{Baselines: }

We calculate the mean predictor (MP) on the train set and
a multi-layer perceptron (MLP) with $\tanh$ activation functions and five hidden layers
with 50 neurons each. It is trained with batch size $100$ for $5000$ epochs and the
\textsc{Adam} optimizer with a learning rate of 0.001 and $\beta_1 = 0.9$.
A grid search on the learning rate revealed that it is pretty robust in the range $[0.001, 0.0001]$.
We select the best validation model to avoid overfitting.

We compare to \dEQL, a state-of-the-art method from \citet{Sahoo2018} with atomic unit types
$\{\sin, \cos, *, \text{identity}\}$ and division in the final layer. It is sufficient to
use the hyperparameters proposed in  \citet{Sahoo2018} since the datasets on which we
compare have similar properties to the ones in \citet{Sahoo2018}.
% (\eg number of data points, input domain, input dimension, output dimension).
So we applied, \textsc{Adam} optimizer with learning rate $0.001$ and $\epsilon=10^{-4}$,
mini-batch size $20$, domain penalty $\eta=10$ with $B=10$,
and 10 atomic units per type in each layer.
The number of total epochs is given by $T=(L-1)\cdot 10,000$, where $L$ is the number of hidden
layers.
Just for the combustion engine dataset, which has about 5 to 6 times fewer data points,
we had to increase the number of total epochs to $T=(L-1)\cdot80,000$.
We perform model selection amongst the following parameters: regularization strengths
$\lambda=10^k$ where $k$ is in the range from $-6.0$ to $-3.5$ with 26 equally spaced steps and
 $L\in\{2,3,4\}$.
This results in 78 equations with different
complexities and root mean squared errors (RMSE).

Another baseline is (\textsc{PySR}, \citet{pysr}) a genetic algorithm for symbolic regression
with hyperparameters shown in \tableref{ta:pysr} for two different configurations (\pysrA, \pysrB).
Both settings turned out to perform well on different datasets.
We had to
restrict the size of the datasets to 1000 data points in order to avoid exploding memory size.
We do not compare to \textsc{Eureqa}, the current state-of-the-art tool for symbolic regression \citep{Dub2011},
since it has become proprietary and was merged into an online service.
\begin{table}[hbt!]
\centering
    \caption{Tuned hyperparameters for the genetic algorithm \citep{pysr}.
    }
\label{ta:pysr}
    \begin{tabular}{c  c c}
    \toprule
        & \pysrA & \pysrB\\
    \midrule
        niterations& 40 & 10\\
        npop & 1000 & 1000\\
        populations& 60& 30\\
        binary operators& $\{\pm, *, /\}$& $\{\pm, *, /\}$\\
        unary operators&$ \{\cos,\exp,\log\}$&$ \{\cos,x^2,\exp,\log\}$\\
        maxsize & 40 & 40\\
        parsimony&0& 0\\
        warmupMaxsizeBy& 0.5& 0\\
        useFrequency& False& True\\
        annealing& False& False\\
        optimizer algorithm& BFGS&BFGS\\
        optimizer iterations&100& 100\\
        procs & 30&30\\
    \bottomrule
    \end{tabular}
\end{table}

For the real world datasets a Gaussian Process (GP) is calculated with \mbox{ASCMO}~\citep{Ascmo2015},
a standard tool from the engineering domain.
\subsection{Power Loss of an Electric Machine}\label{app:power_loss_data}
    Here, we present details on the preparation of training and test dataset.
    First, $20\%$ of the whole dataset is used for testing and excluded from training..%Georg: clarify
    Then, the train domain is restricted to $80\%$ of its range of operation (\([a,b]\rightarrow [0.8 a, 0.8 b]\) 
    for each dimension).
    Rotor temperature $T_{\text{rot}}$ %\matthias[]{in brackets or not? currently not consistent}
    is not restricted, since it consists just of three different operation points.
    Further, $10\%$ of the train dataset is used for validation.
    Restricting the train domain assures that the test dataset contains samples from an extrapolation domain 
    as well as samples from the train domain.

\subsection{Sparsity Inducing Method}\label{app:l0_reg}
In this section we outline the $L_0$ regularization by \citet{Louizos2018}
in more detail, which is used in our sparsity inducing method.
For simplicity the complexity factors $c_u$ are neglected, since it is straightforward to integrate them.
The objective function we want to optimize is given by
\begin{align}
    \Ls = \frac{1}{N} \sum\limits_{i=1}^N
    \norm{\vy_i-\mathrm{iEQL}(\vx_i, \mW)}_2^2 + \lambda \norm{W}_0 \,.
\end{align}
The $L_0$-norm counts the number of non-zero weights, which total number is denoted by $\abs{W}$.
Specific choices of the scale parameter $\lambda$ refer to well-known model selection criteria as
the Bayesian Information Criterion (BIC, \citet{Schwarz1978})  and the Akaike Information
Criterion (AIC, \citet{Akaike1998}).
Each weight of the \bEQL is multiplied by a non-negative stochastic Bernoulli distributed gate
\begin{align}
    w_i = \tilde w_i \cdot g_j &&
    q(g_j\g\pi_j) =\Ber(\pi_j)\,.
\end{align}
The dropout rate of each weight is thus given by $1-\pi_j$.
The expectation of a weight being relevant is given by $\pi_j$.
Those gates learn collectively which weights are relevant.
The expected objective is then
\begin{align}
    \Ls =\E_{q(\vg\g\pi)}\left[\frac{1}{N}\sum\limits_{i=1}^{N} \norm{\vy_i-\mathrm{iEQL}(\vx_i,\mW)}^2_2\right]
    +\lambda\sum\limits_{j\leq\abs{W}}\pi_j\,.
\end{align}
This objective can be described as a special case of a variational bound over weights with
spike and slab \citep{Mitchell1988} priors and approximate posteriors.
We refer the interested reader to appendix A of \citet{Louizos2018} for further details.

\subsubsection{Reparameterization Trick}
In order to apply gradient based optimization the objective is smoothed.
Therefore, the gates $z$ are hard-sigmoid rectifications $b(\cdot)$ of a continuous
random variable $s$ with variables $\phi$
\begin{align}
    s \sim  q(s\g\phi)\,, && z = b(s)\,,&& \text{with: }  b(\cdot)=\min(1, \max(0,\cdot)).
\end{align}
The expected loss is then
\begin{align}
    \Ls\!&=\!\E_{q(s\g\phi)}\!\left[\!\frac{1}{N}\!\sum\limits_{i=1}^{N}
        \norm{\vy_i\!-\!\mathrm{iEQL}\!(\vx_i; \mW\!\odot\! b(s)}^2_2\!\right]\!+\!
        \lambda\!\sum\limits_{j\leq\abs{W}}\! (1\!-\!Q(s_j\!\leq\!0\!\g\!\phi_j))
    % b(\cdot) &= \min(1, \max(0,\cdot))\,,\nonumber
\end{align}
with the cumulative distribution (CDF) $Q(\cdot)$ of $s$.
This penalizes the probability of a gate being non-zero.
By choosing a suitable continuous distribution $q(s)$ we can apply the reparameterization trick with
parameter free noise distribution $\epsilon$ and a deterministic and differentiable transformation
$f(\cdot)$

\begin{align}
    \Ls\!&=\!\E_{p(\epsilon)}\!\left[\!\frac{1}{N}\!\sum\limits_{i=1}^{N}
        \norm{\vy_i\!-\!\mathrm{iEQL}\!(\vx_i; \mW\!\odot\! b(f(\phi,\epsilon))}^2_2\!\right]\!+\!
        \lambda\!\sum\limits_{j\leq\abs{W}}\! (1\!-\!Q(s_j\!\leq\!0\!\g\!\phi_j))
\end{align}
The final reparameterization is given by the hard concrete distribution, which is outlined in the following.

\subsubsection{Hard Concrete Distribution}
$s$ is a binary concrete random variable (\citet{Maddison2016, Jang2016})
by choice, distributed over $(0,1)$ with probability density
$q(s\g\phi)$ and cumulative density $Q_{\beta}(s\g \phi)$.
The distribution has two parameters.
Its location is denoted
by $\log\alpha$ and the degree of approximation is controlled by $\beta$.
A sampling method for the
stretched version for the interval $(\gamma,\zeta)$ is given by
\begin{align}
    u &\sim U(0,1)\\
    s &= \mathrm{Sigmoid}\big((\log u - \log(1-u) + \log \alpha)/\beta\big)\label{eq:s_concrete}\\
    \bar s &= s(\zeta - \gamma) + \gamma\\
    z &= b(\bar s)\label{eq:s_hardconcrete}
\end{align}
with more details
\begin{align}
    q_s(s\g\phi)&=\frac{\beta\alpha s^{-\beta-1}(1-s)^{-\beta-1}}
    {(\alpha s^{-\beta}+ (1-s)^{-\beta/2})^2}\\
    Q_s(s\g\phi) &= \mathrm{Sigmoid}\left((\log s - \log(1-s))\beta - \log\alpha\right)
    \label{eq:concrete}\\
    q_{\bar s}(\bar s\g\phi) &= \frac{1}{\abs{\zeta-\gamma}}
        q_s\left(\frac{\bar s -\gamma}{\zeta-\gamma}\g \phi\right)\\
    Q_{\bar s}(\bar s\g \phi) &=Q_s \left(\frac{\bar s -\gamma}{\zeta-\gamma}\g \phi\right)
    \label{eq:hardconcrete}
\end{align}
and finally the hard sigmoid $z$ we obtain the distribution
\begin{align}
    q(z\g\phi) =& Q_{\bar s}(0\g\phi)\delta(z) + \big(1-Q_{\bar s}(1\g\phi)\big)
        \delta(z-1) \nonumber\\
        &+ \big(Q_{\bar s}(1\g\phi)-Q_{\bar s}(0\g\phi)\big)\,
        q_{\bar s}(z\g\bar s \in (0,1), \phi)
\end{align}
Thus the complexity loss is given by
\begin{align}
    \Ls_{\text{C}} &= \mathrm{Sigmoid}\left((\log \alpha - \beta\log\frac{-\gamma}{\zeta}\right)
\end{align}
and  for final estimation of the parameters under a hard concrete gate:
\begin{align}
    \hat z = \min\left(1,\max\left(0,\mathrm{Sigmoid}\left(\log\alpha\right)(\zeta-\gamma) + \gamma\right)\right)
\end{align}
% \note{is there really $\cdot\beta$ in \eqref{2} and $/\beta$ in \eqref{1}}
In our implementation we use the same hyperparameters as \citet{Louizos2018}
\begin{align}
    \zeta = 1.1, &&
    \gamma = -0.1,&&
    \beta = 2/3\,. \label{eq:hyper}
\end{align}

\begin{figure}[hbt!]
    \subfigure[]  {\small% This file was created by tikzplotlib v0.9.1.
\begin{tikzpicture}

\definecolor{color0}{rgb}{0.12156862745098,0.466666666666667,0.705882352941177}
\definecolor{color1}{rgb}{1,0.498039215686275,0.0549019607843137}

\begin{axis}[
height=0.5\figheight,
legend cell align={left},
legend style={fill opacity=0.5, draw opacity=1, text opacity=1, at={(0.5,0.91)}, anchor=north, draw=white!80!black},
tick pos=left,
width=0.5\figwidth,
x grid style={white!50.1960784313725!black},
xlabel={\(\displaystyle z\)},
xmin=-0.04989, xmax=1.04989,
xtick style={color=black},
xtick={-0.2,0,0.2,0.4,0.6,0.8,1,1.2},
xticklabels={\(\displaystyle -0.2\),\(\displaystyle 0.0\),\(\displaystyle 0.2\),\(\displaystyle 0.4\),\(\displaystyle 0.6\),\(\displaystyle 0.8\),\(\displaystyle 1.0\),\(\displaystyle 1.2\)},
y grid style={white!50.1960784313725!black},
ylabel={\(\displaystyle p(z)\)},
ymin=-0.1, ymax=3,
ytick style={color=black},
ytick={-1,0,1,2,3},
yticklabels={\(\displaystyle -1\),\(\displaystyle 0\),\(\displaystyle 1\),\(\displaystyle 2\),\(\displaystyle 3\)}
]
\addplot [line width=1.4pt, color0]
table {%
0.0001 14.3035892618914
0.0101989898989899 2.85074098308389
0.0202979797979798 2.18658406888592
0.0303969696969697 1.86056568534196
0.0404959595959596 1.65441575768831
0.0505949494949495 1.50805028070756
0.0606939393939394 1.39690093224339
0.0707929292929293 1.30869031119729
0.0808919191919192 1.23647001011692
0.0909909090909091 1.17595643245255
0.101089898989899 1.12433805480575
0.111188888888889 1.07967886462445
0.121287878787879 1.04059416424614
0.131386868686869 1.00606261296596
0.141485858585859 0.97531144120051
0.151584848484848 0.947743290403799
0.161683838383838 0.922887884874713
0.171782828282828 0.900369120961448
0.181881818181818 0.879882060577657
0.191980808080808 0.861176477659648
0.202079797979798 0.844044853224415
0.212178787878788 0.828313459704236
0.222277777777778 0.813835634222881
0.232376767676768 0.800486631051009
0.242475757575758 0.788159631948339
0.252574747474747 0.77676261803574
0.262673737373737 0.766215891303556
0.272772727272727 0.75645009199193
0.282871717171717 0.747404598741072
0.292970707070707 0.73902622727999
0.303069696969697 0.731268164203459
0.313168686868687 0.724089087535822
0.323267676767677 0.717452436953512
0.333366666666667 0.711325804869493
0.343465656565657 0.705680425858664
0.353564646464646 0.70049074667609
0.363663636363636 0.695734062782551
0.373762626262626 0.69139021012669
0.383861616161616 0.687441303145034
0.393960606060606 0.683871511680982
0.404059595959596 0.680666870902707
0.414158585858586 0.67781511940135
0.424257575757576 0.675305561537332
0.434356565656566 0.673128950822104
0.444455555555556 0.671277391711551
0.454554545454545 0.669744257674003
0.464653535353535 0.668524123802526
0.474752525252525 0.667612712585256
0.484851515151515 0.667006851743129
0.494950505050505 0.666704443302742
0.505049494949495 0.666704443302742
0.515148484848485 0.667006851743129
0.525247474747475 0.667612712585256
0.535346464646465 0.668524123802526
0.545445454545455 0.669744257674003
0.555544444444444 0.671277391711551
0.565643434343434 0.673128950822104
0.575742424242424 0.675305561537332
0.585841414141414 0.67781511940135
0.595940404040404 0.680666870902707
0.606039393939394 0.683871511680982
0.616138383838384 0.687441303145034
0.626237373737374 0.69139021012669
0.636336363636364 0.695734062782551
0.646435353535354 0.70049074667609
0.656534343434343 0.705680425858664
0.666633333333333 0.711325804869493
0.676732323232323 0.717452436953512
0.686831313131313 0.724089087535822
0.696930303030303 0.731268164203459
0.707029292929293 0.73902622727999
0.717128282828283 0.747404598741072
0.727227272727273 0.75645009199193
0.737326262626263 0.766215891303556
0.747425252525252 0.776762618035739
0.757524242424242 0.788159631948339
0.767623232323232 0.800486631051009
0.777722222222222 0.813835634222881
0.787821212121212 0.828313459704236
0.797920202020202 0.844044853224415
0.808019191919192 0.861176477659648
0.818118181818182 0.879882060577656
0.828217171717172 0.900369120961448
0.838316161616162 0.922887884874713
0.848415151515151 0.947743290403798
0.858514141414141 0.97531144120051
0.868613131313131 1.00606261296596
0.878712121212121 1.04059416424614
0.888811111111111 1.07967886462445
0.898910101010101 1.12433805480575
0.909009090909091 1.17595643245255
0.919108080808081 1.23647001011692
0.929207070707071 1.30869031119729
0.939306060606061 1.39690093224339
0.949405050505051 1.50805028070756
0.95950404040404 1.65441575768831
0.96960303030303 1.86056568534196
0.97970202020202 2.18658406888592
0.98980101010101 2.85074098308389
0.9999 14.3035892618919
};
\addlegendentry{concrete}
\addplot [line width=1.4pt, color1]
table {%
0.0001 1.01697762838838
0.0101989898989899 0.976122080635324
0.0202979797979798 0.940292820373217
0.0303969696969697 0.908561909229335
0.0404959595959596 0.880228472553066
0.0505949494949495 0.854751439240197
0.0606939393939394 0.831705251421845
0.0707929292929293 0.810749783764566
0.0808919191919192 0.791609362939645
0.0909909090909091 0.774057791922743
0.101089898989899 0.757907441326024
0.111188888888889 0.74300115926905
0.121287878787879 0.729206174748954
0.131386868686869 0.716409436869672
0.141485858585859 0.704514005339353
0.151584848484848 0.693436222127238
0.161683838383838 0.683103471426832
0.171782828282828 0.673452388153741
0.181881818181818 0.664427412281914
0.191980808080808 0.655979612609616
0.202079797979798 0.648065722444212
0.212178787878788 0.640647343454141
0.222277777777778 0.633690284073125
0.232376767676768 0.627164006391503
0.242475757575758 0.621041161150223
0.252574747474747 0.615297194768009
0.262673737373737 0.609910015639035
0.272772727272727 0.604859709493896
0.282871717171717 0.600128295606951
0.292970707070707 0.595699517194765
0.303069696969697 0.591558660584245
0.313168686868687 0.587692398710474
0.323267676767677 0.584088655289755
0.333366666666667 0.580736486645907
0.343465656565657 0.577625978680135
0.353564646464646 0.574748156892043
0.363663636363636 0.572094907701018
0.373762626262626 0.569658909598512
0.383861616161616 0.567433572894546
0.393960606060606 0.565412987015493
0.404059595959596 0.56359187447228
0.414158585858586 0.561965550754493
0.424257575757576 0.560529889521377
0.434356565656566 0.559281292559047
0.444455555555556 0.558216664057715
0.454554545454545 0.557333388835817
0.464653535353535 0.556629314201755
0.474752525252525 0.556102735200378
0.484851515151515 0.555752383041729
0.494950505050505 0.555577416555316
0.505049494949495 0.555577416555316
0.515148484848485 0.555752383041729
0.525247474747475 0.556102735200378
0.535346464646465 0.556629314201755
0.545445454545455 0.557333388835817
0.555544444444444 0.558216664057715
0.565643434343434 0.559281292559047
0.575742424242424 0.560529889521377
0.585841414141414 0.561965550754493
0.595940404040404 0.563591874472279
0.606039393939394 0.565412987015493
0.616138383838384 0.567433572894546
0.626237373737374 0.569658909598513
0.636336363636364 0.572094907701018
0.646435353535354 0.574748156892043
0.656534343434343 0.577625978680135
0.666633333333333 0.580736486645907
0.676732323232323 0.584088655289755
0.686831313131313 0.587692398710474
0.696930303030303 0.591558660584245
0.707029292929293 0.595699517194765
0.717128282828283 0.600128295606951
0.727227272727273 0.604859709493896
0.737326262626263 0.609910015639035
0.747425252525252 0.615297194768009
0.757524242424242 0.621041161150223
0.767623232323232 0.627164006391503
0.777722222222222 0.633690284073125
0.787821212121212 0.640647343454141
0.797920202020202 0.648065722444212
0.808019191919192 0.655979612609616
0.818118181818182 0.664427412281913
0.828217171717172 0.67345238815374
0.838316161616162 0.683103471426832
0.848415151515151 0.693436222127238
0.858514141414141 0.704514005339353
0.868613131313131 0.716409436869671
0.878712121212121 0.729206174748954
0.888811111111111 0.74300115926905
0.898910101010101 0.757907441326024
0.909009090909091 0.774057791922743
0.919108080808081 0.791609362939645
0.929207070707071 0.810749783764566
0.939306060606061 0.831705251421844
0.949405050505051 0.854751439240197
0.95950404040404 0.880228472553066
0.96960303030303 0.908561909229334
0.97970202020202 0.940292820373216
0.98980101010101 0.976122080635323
0.9999 1.01697762838838
};
\addlegendentry{hard concrete}
\end{axis}

\end{tikzpicture}}
    \subfigure[]  {\small\input{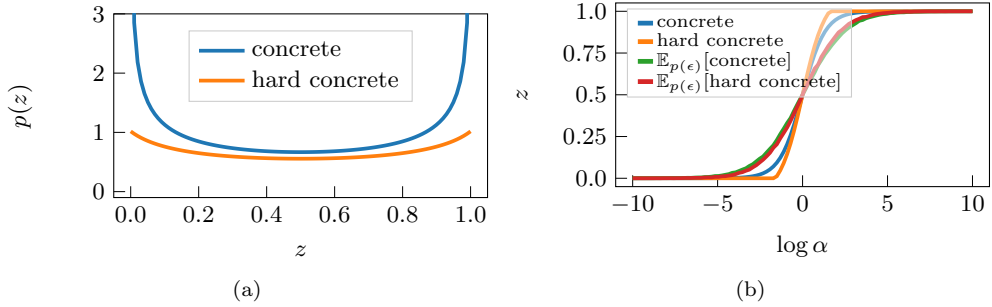}}
    % \input{./fig/l0_F_plot}
    % \end{small}
    \vspace{-.8em}
    \caption{The probability density function for the concrete gate (\eqref{eq:concrete}) and hard concrete gate (\eqref{eq:hardconcrete}) are shown in (a)
    and their expectation (\eqref{eq:s_concrete} and \eqref{eq:s_hardconcrete} respectively) for $10,000$
    samples are shown in (b). We added the concrete and hard concreate without noise as well.
    All plots are produced with the hyperparameters mentioned in \eqref{eq:hyper}.
 		}
        \label{fig:l0_plots}
\end{figure}

\subsection{Model Selection Criteria}\label{app:model_selection}
Both criteria \ValIE and \ValISparse perform similarly good on equations (S0--S6), but
the \ValIE methods performs better on the more sophisticated equations (A1--A4).
However, it requires additional extrapolation data points, which might not be available in real world datasets.
Therefore, we apply the \ValISparse method on the two real world datasets.

\begin{figure}[hbt!]
\centering

    \input{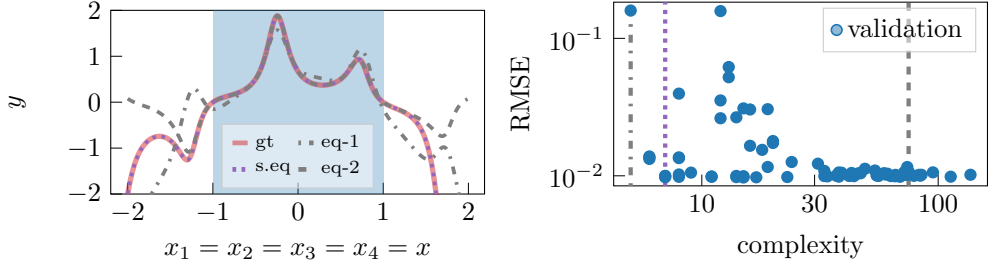}
    \hfill
    % This file was created by tikzplotlib v0.9.1.
\begin{tikzpicture}

\definecolor{color0}{rgb}{0.12156862745098,0.466666666666667,0.705882352941177}
\definecolor{color1}{rgb}{0.580392156862745,0.403921568627451,0.741176470588235}

\begin{axis}[
height=0.5\figheight,
width=0.5\figwidth,
legend cell align={left},
legend style={fill opacity=0.5, draw opacity=1, text opacity=1, at={(0.97,0.97)}, anchor=north east, draw=white!80!black},
log basis x={10},
log basis y={10},
tick pos=left,
x grid style={white!50.1960784313725!black},
xlabel={complexity},
xmin=4.23723427071471, xmax=161.662055066041,
xmode=log,
xtick style={color=black},
xtick={0.1,1,10,30,100,1000,10000},
xticklabels={\(\displaystyle 10^{-1}\),\(\displaystyle 10^{0}\),\(\displaystyle 10\),\(\displaystyle 30\),\(\displaystyle 100\),\(\displaystyle 10^{3}\),\(\displaystyle 10^{4}\)},
y grid style={white!50.1960784313725!black},
ylabel={RMSE},
ymin=0.00845128555965475, ymax=0.183378769694462,
ymode=log,
ytick style={color=black},
ytick={0.0001,0.001,0.01,0.1,1,10},
yticklabels={\(\displaystyle 10^{-4}\),\(\displaystyle 10^{-3}\),\(\displaystyle 10^{-2}\),\(\displaystyle 10^{-1}\),\(\displaystyle 10^{0}\),\(\displaystyle 10^{1}\)}
]
\addplot [semithick, color0, mark=*, mark size=2, mark options={solid}, only marks]
table {%
5 0.159441879675873
6 0.0138266061296205
6 0.013269802903468
6 0.013594647513555
7 0.00999561165097644
7 0.00976351133360703
7 0.00995673768945315
8 0.013452163080114
8 0.039624592982046
8 0.0102310945079886
8 0.00983128082441225
8 0.0136209186477609
9 0.0105492768819925
11 0.00989584782143419
11 0.00982929177912678
12 0.0353539230022265
12 0.157942341937473
12 0.0262299117642005
13 0.0521450585974389
13 0.061873159766527
14 0.0266440084831429
14 0.00998502898686191
14 0.00978285680619793
15 0.00972007073308847
15 0.00986210067689864
15 0.00993139650200481
15 0.0309415313955116
16 0.0303762088160304
16 0.0165364944778007
17 0.00975326640616477
18 0.0154464711580518
19 0.011594456188462
19 0.0305342699712599
20 0.0173385406201268
20 0.017904406386896
23 0.00979740709836369
24 0.012602284453642
31 0.0122489050021799
33 0.0108599666307368
33 0.00995606939250165
34 0.00990941389199387
37 0.0100316369299226
38 0.0098606992873516
38 0.0105740393833721
38 0.0100914625507818
39 0.0102722959464385
39 0.0105372022053728
42 0.0099670653100544
43 0.0102683485140126
43 0.0104886764135809
44 0.00994136387791625
46 0.0104385606832128
47 0.010428977372429
49 0.0112538658311509
54 0.0109271793974538
55 0.0100629375855744
55 0.0100641562997172
56 0.00997951332229659
59 0.0104490954400746
60 0.0102103310166309
61 0.0103637518730957
63 0.00995842139872346
65 0.00995359290249319
66 0.00992544172506462
66 0.0101615871564337
69 0.00983816146686033
72 0.0106872834837604
73 0.0101373358356757
74 0.0115871073115551
75 0.010245936793566
75 0.0107164288678883
79 0.00999339490922283
83 0.0101672054770747
84 0.0100132108667428
85 0.0101710814607427
95 0.0105962750944114
113 0.00985139262501417
137 0.0101756265925042
};
\addlegendentry{validation}
\addplot [ultra thick, white!49.8039215686275!black, dash pattern=on 1pt off 3pt on 3pt off 3pt, forget plot]
table {%
5 0.00845128555965475
5 0.183378769694463
};
\addplot [ultra thick, color1, dotted, forget plot]
table {%
7 0.00845128555965475
7 0.183378769694463
};
\addplot [ultra thick, white!49.8039215686275!black, dashed, forget plot]
table {%
75 0.00845128555965475
75 0.183378769694463
};
\end{axis}

\end{tikzpicture}
\vspace{-1em}
\caption{%
    The panel in the left column displays a slice through the input space of equation S0.
  The blue shading indicates the train domain.% $[-1,1]^4$.
  The selected equation (dotted line) was determined with the \ValIE selection criteria.
  Additionally, a too simple (dashdotted line) and too complex equation (dashed line) are plotted for comparison.
  The right column shows the corresponding pareto plot.
  Each point corresponds to an independent run of the \bEQL with a different complexity regularization.
  The three equations shown on the left are marked with corresponding vertical lines.
}
\label{fig:f}
\end{figure}

To illustrate the importance of the equation selection process
we evaluate the selected equation alongside with a simpler and a more complex solution
on the S0 dataset, see \figref{fig:f}.
If the selected equation is too complex
its validation loss is at noise level, but it is not able to
extrapolate, as seen \eg in the left panel of \figref{fig:f}.
If the selected equation is too simple
validation and extrapolation loss are not at noise level.
%%%%%%%%%%%%%%%%%%%%%%%%%%%%%%%%%%%%%%%%%%%%%%%%%%%%%%%%%%%%%%%%%%%%%%%%%%%%%%%%

\subsection{Different Combustion engine Equations}\label{app:R2_eq}
We present the equations of
median performance and the equations with the smallest number of parameters
selected with the \ValISparse criteria for \bEQL and \bEQLmotor.
An overview of all four analyzed equations on the combustion engine dataset
is shown in \tableref{ta:ieql_eqs}.
% An overview of all four analysed equations on the combustion engine dataset
% is shown in \tableref{ta:ieql_eqs}.
\begin{table}[hbt!]
\centering
    \caption{Overview of analysed \bEQL expressions on the combustion engine dataset.
    The RMSE on the test dataset with the number of active parameters is shown. }
\label{ta:ieql_eqs}
    \begin{tabular}{c  c  c  c}
    \toprule
        \bEQL & RMSE [Nm]        &  \#parameters & equation\\
    \midrule
        motor simple & 6.16 & 15 & \plaineqref{eq:R2_motor_simplEq}\\
        motor median & 3.17 & 32 & \plaineqref{eq:R2_motor_selEq}\\
        plain simple      & 3.04 & 42 & \plaineqref{eq:R2_simplEq}\\
        plain median      & 2.48 & 79 & \plaineqref{eq:R2_selEq}\\
    \bottomrule
    \end{tabular}
\end{table}
Numbers were rounded to three figures and input and
output dimensions were anonymized.
The motor specific complexity factors lead to equations of
mainly polynomial structure with higher degree than without motor specific complexity factors.
The latter combine polynomials with simple nonlinear units like $\cos,\exp, \log$ or division units.
Using more nonlinear units reduces the degree of the used polynomials.

\paragraph*{\bEQLmotor simple:}
    \begin{itemize}
        \item compact notation of a polynomial of degree 8
    \end{itemize}
    \begin{scriptsize}
    \begin{dmath}\label{eq:R2_motor_simplEq}
        y=2.48 x_{2} + 0.66 \left(0.3 x_{2} + 0.79\right) \left(- 0.8 x_{1} + 1.86 x_{3} - 0.08\right) - 0.52 \left(- 2.13 x_{3} + 0.57 \left(0.05 - 1.19 x_{3}\right)^{2} - 0.62 \left(- 0.89 x_{1} + 0.85 \left(0.46 x_{1} - 0.32\right) \left(1.28 x_{3} + 0.34 x_{5} + 0.71\right) + 0.42\right)^{2} + 0.56\right)^{2} + 0.94
    \end{dmath}
    \end{scriptsize}

\paragraph*{\bEQLmotor median:}
    \begin{itemize}
        \item compact notation of a polynomial of degree 7
        \item $\cos$ units in final layer with previous polynomial terms as arguments.
    \end{itemize}
    \begin{scriptsize}
    \begin{dmath}\label{eq:R2_motor_selEq}
        y=- 0.3 x_{1} + 2.32 x_{2} + 0.27 x_{3} + 0.6 c_2 + 0.3 c_3 + 0.35 \cos{\left(1.12 x_{3} + 1.42 c_2 + 0.59 c_3 - 0.23 \right)} + 0.51

    \end{dmath}
    \end{scriptsize}
substitutions:
    \begin{scriptsize}
% substitutions:
    \begin{align*}
        c_1 =& \left(0.05 - 0.69 x_{1}\right) \left(1.0 x_{1} + 0.64\right)\\
c_2 =& \left(- 0.95 x_{2} - 0.69\right) \left(- 1.06 x_{3} + 0.96 \left(- 0.21 x_{1} + 0.66 x_{3} - 1.01\right)^{2} - 1.2\right)\\
c_3 =& \left(- 0.74 x_{3} - 0.76 \left(0.6 x_{1} + 0.73 c_1 + 0.47 \left(0.76 x_{2} + 0.19 x_{4} - 0.19 x_{5} - 0.6\right)^{2} - 0.9\right)^{2} + 1.36\right)\\
     & \left(2.0 x_{3} + 0.88 \left(- 0.73 x_{1} - 1.17 c_1 + 0.26\right) \left(0.6 x_{4} - 0.72 x_{5} - 0.46\right) - 0.31\right)

    \end{align*}
    \end{scriptsize}

\paragraph*{\bEQL simple:}
    \begin{itemize}
        \item polynomial of degree 4 with simple $\cos$ and $\exp$ units
        \item $\cos$ units with previous polynomial terms as arguments.
    \end{itemize}
    \begin{scriptsize}
    \begin{dmath}\label{eq:R2_simplEq}
        y=1.49 x_{2} + 0.53 \left(0.94 x_{2} + 0.8\right) \left(1.89 x_{3} - 0.29 \cos{\left(4.79 x_{3} + 2.83 \right)} + 0.46\right) + 0.81 e^{- 0.27 x_{1} + 0.18 x_{2}} + 0.94 e^{- 0.32 x_{1} + 0.25 x_{2} + 0.55 x_{3}} - 0.05 c_1+ 0.22 \cos{\left(1.45 x_{3} - 1.02 c_2 + 0.07 \right)} - 0.29 \cos{\left(0.66 x_{2} - 1.13 c_2 - 0.74 c_3 + 2.73 \right)} - 0.23 \cos{\left(- 1.21 x_{1} + 1.99 x_{3} - 2.26 \left(- 0.94 x_{2} - 0.52\right) \left(0.97 x_{3} + 0.02\right) - 0.97 c_3 + 3.71 \right)} - 1.71

    \end{dmath}
    \end{scriptsize}
substitutions:
    \begin{scriptsize}
% substitutions:
    \begin{align*}
        c_1 =&\cos{\left(4.0 x_{1} + 2.38 \right)} \\
c_2 =&\left(- 0.49 x_{1} + 1.1 x_{3} + 0.76 \left(1.02 x_{1} - 0.02\right)^{2} - 0.85\right)^{2}\\
c_3 =& \left(- 0.41 x_{2} - 0.53 x_{3} - 0.29 x_{4} + 0.34 x_{5} + 0.59 \left(0.07 - 0.97 x_{1}\right) \left(0.28 - 1.02 x_{2}\right) + 0.23 c_1+ 0.6\right)^{2}

    \end{align*}
    \end{scriptsize}

\paragraph*{\bEQL median:}
\begin{itemize}
    \item    polynomial of degree 4 with simple $\cos, \log$ and division units
    \item    division units occur also in intermediate layers
    \item    $\cos$ units with previous polynomial terms as arguments
\end{itemize}
    \begin{scriptsize}
    \begin{dmath}\label{eq:R2_selEq}
        y=2.0 x_{2} + 0.6 x_{3} + 0.06 x_{4} - 0.05 x_{5} - 0.49 \left(0.83 x_{2} + 0.58\right) \left(- 0.81 x_{3} - 0.42\right) + \frac{0.39 \left(0.99 x_{2} + 0.79\right)}{6.75 \left(1.77 c_1 - 0.75\right)^{2} + 1.96} - 0.24 \left(- 0.77 x_{2} - 0.36 c_1 - 0.42\right) \left(1.11 x_{3} + 0.32 c_2 - 0.89 \left(- 1.1 x_{3} + 0.73 c_1 + 0.41\right)^{2} - 0.47 \cos{\left(5.98 x_{3} + 3.27 \right)} + 0.09\right) - 0.29 \left(- 0.78 x_{2} + 0.35 \left(1.77 c_1 - 0.75\right)^{2} + 0.35 \cos{\left(1.72 x_{2} - 2.6 \right)} - 0.67\right)^{2} - 0.28 c_3+ 0.18 \log{\left(1.08 - 0.88 x_{1} \right)} - 0.2 \cos{\left(- 1.21 x_{2} + 1.28 c_3+ 4.61 \right)} - 0.15 \cos{\left(1.95 x_{2} + 1.58 x_{3} + 1.0 c_4 + 3.51 \right)} + 0.18 \cos{\left(0.77 x_{1} - 1.71 x_{2} - 0.76 c_4 + 0.71 c_3+ 1.51 \right)} + 0.74 - \frac{0.39 \left(0.85 x_{1} - 0.58\right)}{3.29 - 2.85 x_{2}}

    \end{dmath}
    \end{scriptsize}
substitutions:
    \begin{scriptsize}
    \begin{align*}
        c_1 = &\cos{\left(2.07 x_{1} + 0.21 x_{2} - 0.02 \right)}\\
c_2 = &\left(0.33 x_{1} + 1.09 c_1 + 0.51\right) \left(- 0.74 x_{4} + 0.61 x_{5} + 1.02 c_1 - 0.62\right)\\
c_3 = &\left(0.29 x_{1} - 1.65 x_{3} - 0.09 x_{4} + 0.09 x_{5} - 0.27 \left(- 0.69 x_{1} - 0.14\right)^{2} + 0.35 \left(0.68 x_{1} - 1.27 x_{3} + 0.3\right)^{2} + 0.4\right)^{2} \\
c_4 =&\left(- 0.43 x_{4} + 0.36 x_{5} + 0.8 + \frac{0.86 \left(- 1.22 x_{2} - 0.94 x_{3} - 0.66\right)}{2.11 x_{1} + 2.35}\right)\\
&\left(0.6 x_{2} + 1.28 x_{3} + 0.89 x_{4} - 1.18 x_{5} + 0.77 c_2 - 0.05\right)

    \end{align*}
    \end{scriptsize}

\end{document}